\documentclass[sigconf]{acmart}

\AtBeginDocument{%
  \providecommand\BibTeX{{%
    \normalfont B\kern-0.5em{\scshape i\kern-0.25em b}\kern-0.8em\TeX}}}

\usepackage{subfigure}

\usepackage{multirow}
\usepackage{algorithm}
\usepackage{algorithmic}
\usepackage{graphicx} 
\usepackage{float}

\newcommand*{\Scale}[2][4]{\scalebox{#1}{$#2$}}
\setcopyright{acmcopyright}
\copyrightyear{2018}
\acmYear{2018}
\acmDOI{10.1145/1122445.1122456}

\acmConference[SIGKDD 2022]{SIGKDD 2022}{August 14-18, 2022}{Washington DC}

\begin{document}

\title{Deconfounding Actor-Critic Network with Policy Adaptation for Dynamic Treatment Regimes}

\author{Changchang Yin}
\affiliation{%
 \institution{The Ohio State University}
  \streetaddress{281 W Lane Ave}
  \city{Columbus}
  \state{OH}
  \country{USA}
  \postcode{43210}
}
\email{yin.731@osu.edu}

\author{Ruoqi Liu}
\affiliation{%
 \institution{The Ohio State University}
  \streetaddress{281 W Lane Ave}
  \city{Columbus}
  \state{OH}
  \country{USA}
  \postcode{43210}
}
\email{liu.7324@osu.edu}

\author{Jeffrey Caterino}
\affiliation{%
 \institution{The Ohio State University Wexner Medical Center}
  \streetaddress{281 W Lane Ave}
  \city{Columbus}
  \state{OH}
  \country{USA}
  \postcode{43210}
}
\email{jeffrey.caterino@osumc.edu}

\author{Ping Zhang}
\orcid{0000-0002-4601-0779}
\affiliation{%
 \institution{The Ohio State University}
  \streetaddress{281 W Lane Ave}
  \city{Columbus}
  \state{OH}
  \country{USA}
  \postcode{43210}
}
\email{zhang.10631@osu.edu}

\renewcommand{\shortauthors}{Trovato and Tobin, et al.}

\begin{abstract}
Despite intense efforts in basic and clinical research, an individualized ventilation strategy for critically ill patients remains a major challenge. Recently, dynamic treatment regime (DTR) with reinforcement learning (RL) on electronic health records (EHR) has attracted interest from both the healthcare industry and machine learning research community. However, most learned DTR policies might be biased due to the existence of confounders. Although some treatment actions non-survivors received may be helpful, if confounders cause the mortality, the training of RL models guided by long-term outcomes (e.g., 90-day mortality) would punish those treatment actions causing the learned DTR policies to be suboptimal. In this study, we develop a new deconfounding actor-critic network (DAC) to learn optimal DTR policies for patients. To alleviate confounding issues, we incorporate a patient resampling module and a confounding balance module into our actor-critic framework. To avoid punishing the effective treatment actions non-survivors received, we design a short-term reward to capture patients' immediate health state changes. Combining short-term with long-term rewards could further improve the model performance. Moreover, we introduce a policy adaptation method to successfully transfer the learned model to new-source small-scale datasets. The experimental results on one semi-synthetic and two different real-world datasets show the proposed model outperforms the state-of-the-art models. The proposed model provides individualized treatment decisions for mechanical ventilation that could improve patient outcomes.

\end{abstract}



\maketitle

\section{Introduction}
 
Mechanical ventilation is one of the most widely used interventions in admissions to the intensive care unit (ICU). Around 40\% of patients in the ICU are supported on invasive mechanical ventilation at any given time, accounting for 12\% of total hospital costs in the United States \cite{ambrosino2010difficult,wunsch2013icu}. 
Despite intense efforts in basic and clinical research, an individualized ventilation strategy for critically ill patients remains a major challenge \cite{mechnical_ventilation,mechnical_ventilation_prasad}.
If not applied adequately, suboptimal ventilator settings can result in ventilator-induced lung injury, hemodynamic instability, and toxic effects of oxygen. Dynamic treatment regime (DTR) learning on electronic health records (EHR) with reinforcement learning (RL) might be helpful for learning optimal treatments by analyzing a myriad of (mostly suboptimal) treatment decisions. 

Recently, DTR learning with RL has attracted the interest of healthcare researchers \cite{aiclinician,raghu2017deep,raghu2019reinforcement,ciq,cirl,amia2018,mechnical_ventilation}. However, most existing studies suffer from three limitations. First, most existing RL-based methods \cite{aiclinician,kdd2018,amia2018,mechnical_ventilation} punish the treatment actions for patients who ultimately suffer from mortality. However, for some patients with worse health states, the mortality rates remain high even if they received optimal treatment. Actions that did not contribute to mortality should not be punished in the treatment of non-survivors. Second, RL strategies learned from initial EHR datasets may be biased due to the existence of confounders (patients' health states are confounders for treatment actions and clinical outcomes) and data unbalance (mortality rates in different datasets vary widely and might be less than 25\%). Third, external validation on different-source data is lacking (e.g., how a model trained on data extracted from the United States performs on European datasets). Especially when the treatment action distributions are different, efficient adaptation to new datasets has not been considered.

In this study, we propose a new deconfounding actor-critic model (\textbf{DAC}) to address these issues.
First, we resample paired survivor and non-survivor patients with similar estimated mortality risks to build balanced mini-batches.
Then we adopt an actor-critic model to learn the optimal DTR policies. The longitudinal patients' data are sent to a long short-term memory network (LSTM) \cite{lstm} to generate the health state sequences.
The actor network produces the probabilities of different treatment actions at next time step and is trained by maximizing the rewards generated by the critic network. To avoid punishing some effective treatment actions in EHR history of non-survivors, the critic network produces both short-term and long-term rewards. Short-term rewards can encourage the treatment actions that improve patients' health states at coming time steps, even if the patients ultimately suffer from mortality.
To further remove the confounding bias, we introduce a dynamic inverse probability of treatment weighting method to assign weights to the rewards at each time step for each patient and train the actor network with the weighted rewards.
Finally, we introduce a policy adaptation method to transfer well-learned models to new-source small-scale datasets. The policy adaption method chooses actions so that the resulting next-state distribution on the target environment is similar to the next-state distribution resulting from the recommended action on the source environment.

We conduct DTR learning experiments on a semi-synthetic dataset and two real-world datasets (i.e., MIMIC-III \cite{mimic} and AmsterdamUMCdb \cite{amsterdamumcdb}). The experimental results show that the proposed model outperforms the baselines and can reduce the estimated mortality rates. Moreover, we find the mortality rates are lowest in patients for whom clinicians’ actual treatment actions matched the model's decisions.
 The proposed model can provide individualized treatment decisions that could improve patients' clinical outcomes.

In sum, our contributions are as follows: (i) We develop a new DTR learning framework with RL and experiments on MIMIC-III and AmsterdamUMCdb datasets demonstrate the effectiveness of the proposed model; (ii) We present a patient resampling operation and a confounding balance module to alleviate the confounding bias; (iii) We propose combining long-term and short-term rewards to train the RL models; (iv) We propose a policy adaptation model that can effectively adapt pre-trained models to new small-scale datasets.  The code of our proposed DAC model can be found at GitHub\footnote{\label{github}\url{https://github.com/yinchangchang/DAC}}.


\section{Problem Formulation}
\textbf{Setup.}
DTR is modeled as a Markov decision process (MDP) with finite time steps and a deterministic policy consisting of an action space $\mathcal{A}$, a hidden state space $\mathcal{S}$, a observational state space $\mathcal{O}$, and a reward function: $\mathcal{A} \times \mathcal{S} \rightarrow R$. 
A patient's EHR data consists of a sequence of observational variables (including demographics, vital signs and lab values), denoted by $O=\{o_1, o_2, ..., o_T\}$, $o_t \in \mathcal{O}$, the treatment actions represented as $A=\{a_1, a_2, ..., a_T\}$, $a_t \in \mathcal{A}$ and mortality outcome $y \in \{0, 1\}$, where $T$ denotes the length of the patient's EHR history.
We assume some hidden variables $S=\{s_1, s_2, ..., s_T\}$, $s_t \in \mathcal{S}$ can represent the health states of a patient and include the key information of previous observational data of the patients. 
Given the previous hidden state sequence $S_t = \{s_1, s_2, ..., s_t\}$, action sequence $A_{t-1} = \{a_1, a_2, ..., a_{t-1}\}$ and observation sequence $O_t = \{o_1, o_2, ..., o_t\}$ up to time step $t$, our goal is to learn a policy $\pi_\theta(\cdot|S_t, O_t, A_{t-1})$ to select the optimal action $\hat{a}_t$ by maximizing the the sum of discounted rewards (return) from time step $t$. We use LSTM to model patient health states and  LSTM can remember the key information of patients' EHR history. We assume state $s_t$ contains the key information of the previous data, and learn a policy $\pi_\theta(\cdot|s_t)$ instead of $\pi_\theta(\cdot|S_t, O_t, A_{t-1})$.

\begin{figure*}[!t]
\centering  
\includegraphics[width=0.85\textwidth]{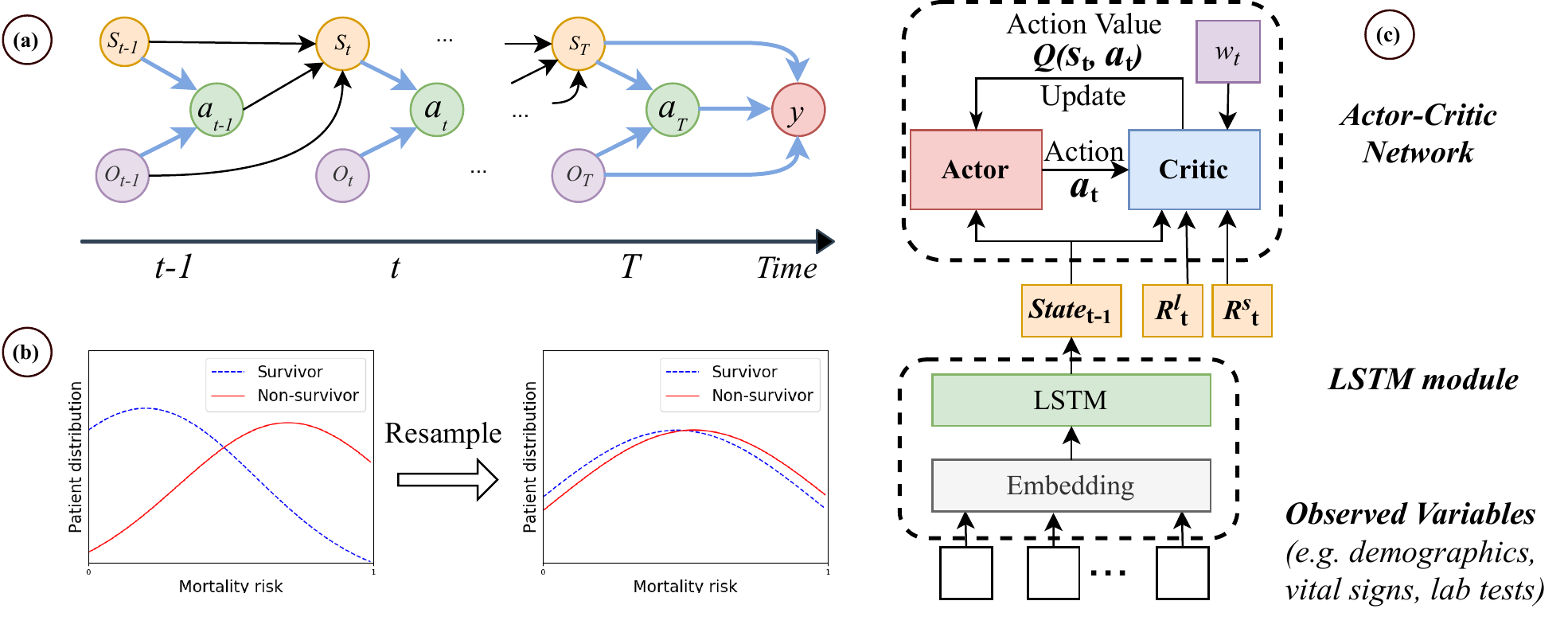} 
\caption{\textbf{Framework of proposed DTR learning model}. \textbf{(a)} The causal graph of variables. $a_t$ denotes the assigned actions. The observational variables $o_t$ are covariates. $y$ denotes the final clinical outcomes. The patient health states $s_t$ are hidden confounders for both $a_t$ and $y$. 
\textbf{(b)} Patient resampling operation. Non-survivors have more high-risk health states than survivors. The unbalanced data might introduce bias to learned DTR policies. We resample the patients according to their mortality risks such that both survivor and non-survivor groups follow similar mortality risk distributions.  \textbf{(c)} Framework of the proposed model.
Given the resampled datasets, the embeddings of observed variable $o_t$ are sent to LSTM to model the patients' health state sequences. Actor network generates the probabilities for next actions based on the health states and critic network produces the short-term reward $R^s_t$ and long-term reward $R^l_t$ for the ($s_t, a_t$) pairs. Considering the causal relationship among states $s_t$, observations $o_t$, actions $a_t$ and outcome $y$, we compute an inverse weight $w_t$ at each time step $t$ for the rewards. 
The actor network is trained by maximizing the expected weighted reward.} 
\label{fig:model} 
\end{figure*}

\vspace{5pt}
\noindent
\textbf{Time-varying confounders.}
Figure \ref{fig:model} (a) shows the causal relationship of various variables.  $o_t$ denotes the time-dependent covariates of the observational data at time step $t$, which is only affected by hidden state $s_t$. The treatment actions $a_t$ are affected by both observed variable $o_t$ and hidden state $s_t$. The potential outcomes $y$ are affected by last observational variable $o_T$, treatment assignments $a_T$ and hidden state $s_T$. Patients' health states $S$ are time-varying confounders for both treatment actions $A$ and clinical outcomes $y$. Without the consideration of the causal relationship among the variables, it is possible that RL models may focus on the strong correlation between positive outcomes and "safe" actions (e.g., without mechanical ventilator) and prefer to recommend the "safe" actions, which will cause much higher mortality rates for high-risk patients. It is very important to remove the confounding when training DTR policies on real-world datasets. 
DTR policies learned from initial clinical data could be biased due to the existence of time-varying confounders.

We summarize the important notations in this paper in Table \ref{tab:notations}.

\begin{table}[!h]
\caption{Important Notations}
\label{tab:notations}
\Scale[0.9]{
{\renewcommand{\arraystretch}{1.3}%
\begin{tabular}{ll}\hline
Notation    &   Definition  \\\hline
$\mathcal{O}$     &   The space of time-varying covariates \\ 
$\mathcal{A}$   &   The set of treatment options of interest   \\  
$\mathcal{S}$   &   The space of hidden confounders \\
$o_{t}$   &   The time-varying covariates at time $t$  \\ 
$a_{t}$   &   The treatment assigned at time $t$  \\
$s_{t}$   &   The hidden state at time $t$  \\
$w_t$ & The reward weight at time $t$ \\
$y$    &   The outcome \\   
$\pi_\theta$ & The learned DTR policy \\
$\rho$ & The state distribution \\
$R^l$ & The long-term reward \\
$R^s$ & The short-term reward \\
$Q$ & The reward for treatment actions \\
$p^m$ & The patient mortality probability \\
$\alpha$ &  The hyper-parameter to adjust the weights of two rewards \\
$w_*, b_*$ & The learnable parameters \\
\hline
\end{tabular}
}\quad }
\vspace{-10pt}
\end{table}

\section{Method}

In this section, we propose a new causal reinforcement learning framework to learn optimal treatment strategies. We first introduce the deconfounding module that resamples patients according to their mortality risks and computes the weights for rewards in RL model. Then we develop an actor-critic network to learn DTR policies with the weighted rewards. Finally, we present a policy adaptation method that can transfer well-trained models to new-source environments.

\subsection{Deconfounding Module}

DTR policies learned from initial clinical data could be biased for two-fold reasons. First, the training of RL models is usually guided by designed rewards, which are highly related to patients' long-term outcomes. Existing DTR models \cite{aiclinician,kdd2018,raghu2017deep} encourage the treatment actions that survivors received and punish the treatment actions that non-survivors received. The mortality rates of the collected datasets have important effects on the learned policies and might cause policy bias. The mortality rates of different-source datasets vary widely and the bias could further limit the model performance when adapting learned DTR policies to new-source datasets. The second reason for policy bias is the existence of confounders. Patients' clinical outcomes $y$ (e.g., mortality or discharged) are affected by both patient health states $s_t$ and treatment actions $o_t$, as shown in Fig. \ref{fig:model} (a). The treatment actions are also affected by patient health states $s_t$. The patient hidden states $s_{t}$ are confounders for both actions $a_{t}$ and final clinical outcome $y$. 
In this subsection, we introduce patient resampling module and confounding balance module to address the policy bias problems.


\vspace{5pt}
\noindent
\textbf{Patient resampling module.} 
We resample the patients according to their mortality risks when training our treatment learning models. First, we train a mortality risk prediction model, which takes the patients' observational data as inputs and produces the 90-day mortality probability at each time step $t$.
Then, patients are divided into a survivor pool and a non-survivor pool. When training treatment learning models, we always sample paired patients from the two pools respectively with similar maximal mortality risks in their EHR sequence. With the resampling operation, we build balanced mini-batch where survivors and non-survivors have similar mortality risk distributions, as shown in Figure \ref{fig:model} (b).

\vspace{5pt}
\noindent
\textbf{Confounding balance module.}
To adjust the confounder, we train the actor-critic network with weighted rewards and the weights are computed based on the probabilities that the corresponding treatment actions are assigned. Given a patient health state $s_t$ at time step $t$, the probability that an action $a$ would be assigned is represented as $\pi_\theta(a|s_t)$. We compute the weights using inverse probability of treatment weighting (IPTW) \cite{iptw,iptwm} and extend to dynamic multi-action setting as follows,  
\begin{equation} 
\label{eq:weight}
w_t = \Pi^t_{\tau=1} \frac{f(a_\tau|A_{\tau - 1})}{f(a_\tau|A_{\tau-1},O_{\tau -1})}
=\Pi^t_{\tau=1} \frac{f(a_\tau|A_{\tau - 1})}{\pi^c(a_\tau|s_\tau)}
\end{equation}
where $f(a_\tau|A_{\tau - 1})$ is the posterior probability of action $a_\tau$ given last action sequence $A_{\tau - 1}$, which could be modelled with LSTM. $f(a_\tau|A_{\tau-1},O_{\tau})$ denotes predicted probability of receiving treatment $a_\tau$
given the observed data and historical information, and is computed with clinician policy $\pi^c$. $\pi^c(a_\tau|s_\tau)$ is the probability for action $a_\tau$ given patient's health state $s_\tau$. $\pi^c$ shares the same actor network as the proposed DAC model and is trained by mimicking clinicians' policy.
The computed weights are used in the training of the actor network.

\subsection{Actor-Critic Framework}

In this subsection, we present the details of our RL model based on actor-critic network, including how to model patients' health states and update the actor and critic networks.

\vspace{5pt}
\noindent
\textbf{Observational data embedding and health state representation.} 
The observational data contain different vital signs and lab tests, which have lots of missing values. Existing models usually impute the missing values based on previous observational data. However, for some patients with some high missing-rate variables, the imputation results might be inaccurate and thus introduce more imputation bias, which is harmful for modeling the patient health states. Following \cite{tame}, we embed the observed variables with corresponding values, and only input the embeddings of observed variables to the model. Given the variable $i$  and the observed values in the whole dataset, we sort the values and discretize the values into $V$ sub-ranges with equal number of observed values in each sub-range. The variable $i$ is embedded into a vector $e^i \in R^k$ with an embedding layer. As for the sub-range $v (1 \leq v \leq V)$, we embed it into a vector $e'^v \in R^{2k}$: 
\vspace{-1mm}
\begin{equation} 
\label{eq:value-embedding-1} 
e'^v_j = sin(\frac{v*j}{V*k}), \qquad
e'^v_{k+j} = cos(\frac{v*j}{V*k}), 
\end{equation}
where $0 \leq j < k$. By concatenating $e^i$ and $e'^v$, we obtain vector containing both the variable's and its value's information. A fully connected layer is followed to map the concatenation vector into a new value embedding vector $e^{iv} \in R^k$. 
 
Given the value embeddings of observational variables in the same collection, a max-pooling layer is followed to generate the collection representation vector $e_t$.
They are sent to a LSTM to generate a sequence of health state vectors  $S=\{s_1, s_2, ..., s_{|S|}\}$, $s_t \in R^k$.

\vspace{5pt}
\noindent
\textbf{Actor network update.}
Given a patient's health states, a fully connected layer and a softmax layer are followed to generate the probabilities for next actions. 
The actor network generates the probabilities for next actions $\pi_\theta$. 
The critic network produce the rewards for action $a$, denoted as $Q(s, a)$. We update the actor network by maximizing the expected reward:
\vspace{-1mm}
\begin{equation}
\label{eq:expected-reward}
J(\pi_\theta) = \int_{s \in \mathcal{S}} \rho(s) \sum_{a \in \mathcal{A}} \pi_\theta(a|s) Q(s,a) ds,
\end{equation}
where $\rho(s)$ denotes the state distribution. We use policy gradient to learn
the parameter $\theta$ by the gradient $\triangledown_\theta J(\pi_\theta)$ which is calculated using
the policy gradient theorem \cite{policy_gradient}:
\vspace{-1mm}
\begin{equation}
\label{eq:policy-gradient}
\begin{aligned}
\triangledown_\theta J(\pi_\theta) = \int_{s \in \mathcal{S}} \rho(s) \sum_{a \in \mathcal{A}} \triangledown_\theta \pi_\theta(a|s) Q(s,a) ds \\
= E_{s \sim \rho, a \in \pi_\theta} [\triangledown_\theta \log \pi_\theta(a|s) Q(s,a)]
\end{aligned}
\end{equation}
\vspace{-2mm}

\begin{algorithm}[t]
\caption{Deconfounding Actor-Critic}
\label{alg:RL}

\textbf{Input}: Observations $O$, treatment actions $A$, outcome $y$;

\noindent
\textbf{Output}: Policy $\pi_\theta$;
\begin{algorithmic}[1] 
\STATE Train a mortality risk prediction model and compute the risks for patients in training set;
\REPEAT
\STATE Sample paired patients from survivor and non-survivor pools with similar mortality risks;
\STATE \textit{ \# Inference}
\FOR{$t = 1, ..., T$} 
\STATE Input the observations $o_t$ to LSTM to generate health states $s_t$;
\STATE Produce probability distribution for next actions $\pi_\theta(\cdot|s_t)$;
\STATE Compute reward weight $w_t$ according to Eq. (\ref{eq:weight});
\STATE Compute long-term reward $R^l(s_t, a_t)$ according to Eq. (\ref{eq:long-term-reward});
\STATE Compute short-term reward $R^s(s_t, a_t)$ according to Eq. (\ref{eq:short-reward});
\STATE Compute the weighted reward $Q(s_t, a_t)$ according to Eq. (\ref{eq:weighted-reward});
\ENDFOR
\STATE \textit{ \# Actor network update}
\STATE Update policy  $\pi_\theta$ according to Eq. (\ref{eq:policy-gradient});
\STATE \textit{ \# Critic network update}
\STATE Update long-term reward function $R^l(s,a)$ by minimizing $J(w_l, b_l)$ in Eq. (\ref{eq:long-term-reward-loss});
\STATE Update mortality risk prediction function $p_m(s,a)$ by minimizing $J(w_m, b_m)$ in Eq. (\ref{eq:long-term-reward-loss});
\UNTIL{Convergence.}
\end{algorithmic}
\end{algorithm}

\vspace{5pt}
\noindent
\textbf{Critic network update.} The critic network takes patients' health states and treatment actions as inputs, and output the rewards. 
We use fully connected layers to learn the long-term reward function:
\vspace{-1mm}
\begin{equation}
\label{eq:long-term-reward} 
R^l(s_t) = s_t w_l + b_l,
\end{equation}
where $w_l \in R^{k \times |\mathcal{A}|}$, $b_l \in R^{|\mathcal{A}|}$ are learnable parameters.
Given the state-action pairs at time $t$, the long-term reward function is trained by minimizing $J(w_l, w_l)$:
\vspace{-1mm}
\begin{equation}
\label{eq:long-term-reward-loss}
\begin{aligned}
J(w_l, b_l) = E_{s_t \sim \rho} [ R^l(s_t, a_t) - z_t)^2] \\
z_t = R^m(s_t,a_t) + \gamma R^l(s_{t+1}, \hat{a}_{t+1}),
\end{aligned}
\end{equation}
where $\hat{a}_{t+1}$ is the action with the maximum reward in the next step, $R^l(s_t, a_t) \in R$ is the corresponding dimension reward of $R^l(s_t)$ for action $a_t$  and $R^m(s_t, a_t)$ denotes the reward at the last time step. Given a patient with EHR length equal to $T$, $R^m(s_t, a_t) = 0, t<T$. Following \cite{aiclinician,raghu2019reinforcement}, the reward for the last action is set as $\pm 15$. Specially, if the patient suffers from mortality, $R^m(s_T, a_T) = -15$. Otherwise, $R^m(s_T, a_T) = 15$.

Most existing RL-based models are trained with long-term rewards and punish the actions non-survivors received. However, for some patients with worse health states, the probability of mortality is still high even if they receive optimal treatment. Some actions should not be punished in the treatment of patients with mortality. We propose a short-term reward based on estimated mortality risk to improve the training of RL models. The estimated mortality risks $p_m(s_t)$ are generated with fully connected layers and a Sigmoid layer:
\vspace{-1mm}
\begin{equation}
\label{eq:mortality-risk-prediction}
p_m(s_t) = Sigmoid (s_t w_m + b_m),
\end{equation}
where $w_m \in R^{k \times |\mathcal{A}|}$, $b_m \in R^{|\mathcal{A}|}$ are learnable parameters.
The mortality probability with an action $a$ at time $t$ is the action's corresponding dimension of $p_m(s_t)$, denoted as $p_m(s_t, a)$. The mortality risk prediction function $p_m$ is trained by minimizing $J(w_m, b_m)$:
\vspace{-1mm}
\begin{equation}
\label{eq:mortality-risk-prediction-loss}
J(w_m, b_m) = E_{s_t \sim \rho} [-y\log (p_m(s_t, a_t)) - (1-y) \log (1 - p_m(s_t, a_t))]
\end{equation}
The short-term reward is computed as the mortality probability decrease given the action as follows, 
\begin{equation}
\label{eq:short-reward}
R^s(s_t, a_t) = \sum_{a \in \mathcal{A}} \pi_\theta(a|s_t) p_m(s_t, a) - p_m(s_t, a_t)
\end{equation}

The overall reward $Q$ is computed by combining short-term and long-term reward:
\begin{equation}
\label{eq:weighted-reward}
Q(s_t,a_t) = w_t ( \alpha R^l(s_t,a_t) +  (1-\alpha)R^s(s_t,a_t)),
\end{equation}
where $\alpha$ is a hyper-parameter to adjust the weights of the two rewards and $w_t$ denotes the inverse weight computed by the confounding balance module. The details of $\alpha$ selection can be found in supplementary material. Algorithm \ref{alg:RL} describes the training process of the proposed DAC.

\subsection{Policy adaptation}
In real-world clinical settings, a pre-trained model might suffer from performance decline in new environments when the patient distribution is different. It is possible that we cannot collect enough data to train a new model in the new environment. To address the problem, we propose a policy adaptation method to transfer the pre-trained model to new environments.

We first train a policy $\pi_\theta^S$ on a source dataset (i.e., MIMIC-III), and then adapt the model to a target dataset (i.e., AmsterdamUMCdb). We learn two dynamic function $f^S$ and $f^T$ on the source dataset and the target dataset respectively to predict next state $s_{t+1}$ given the state $s_t$ and action $a_t$ at time step $t$.
\vspace{-1mm}
\begin{equation}
\label{eq:dynamic-function}
f^T(s_t, a_t) = s_t w_d + b_d,
\end{equation}
where $w_d \in R^{k \times |\mathcal{O}|}$, $b_d \in R^{|\mathcal{O}|}$ are learnable parameters. The dynamic functions are trained by minimizing $J(w_d, b_d)$:
\vspace{-1mm}
\begin{equation}
\label{eq:MLE}
\begin{aligned}
J(w_d, b_d) = E_{s_t \sim \rho} [ f^T(s_t, a_t) - s_{t+1})^2]
\end{aligned}
\end{equation} 
Note that $f^S$ and $f^T$ share the same structure and objective function, but are trained on different datasets. The target dynamics $f^T$ is initialized as source dynamics $f^S$ and fine-tuned on the small-scale target dataset.

Given $\pi_\theta^S$, $f^S$ and $f^T$, we define the policy $\pi_\theta^T$ on target dataset as:
\begin{equation}
\pi_\theta^T(s) = \arg \min _{a \in A} ( f^T(s, a)  - f^S(s, \pi_\theta^S(s)))^2
\label{eq:transfer-learning} 
\end{equation}
Assuming $f^S$ and $f^T$ are accurate in terms of modeling patient state transition on source and target environments, $\pi_\theta^T(s)$ can pick the action such that the resulting next state distribution under $f^T$ on target environment is similar to the next state distribution resulting from $\pi_\theta^S(s)$ under the source dynamics. Algorithm \ref{alg:policy-adaptation} describes the details of policy adaptation.

\begin{algorithm}[!tb]
\caption{Policy Adaptation}
\label{alg:policy-adaptation}

\textbf{Input}: Source domain policy $\pi^S_\theta$, source dynamics $f^S$, patient state $s$;

\noindent
\textbf{Output}: Next action on target domain $\pi^T(s)$,  target dynamics $f^T$;
\begin{algorithmic}[1] 
\STATE Initialize $f^T$ = $f^S$;
\STATE \# Learn the target dynamics $f^T$;
\REPEAT
\STATE Sample a bath patients; 
\FOR{$t = 1, ..., T$} 
\STATE Compute $f^T(s_t,a_t)$;
\ENDFOR
\STATE Update $f^T$ by minimizing $J(w_d, b_d)$;
\UNTIL{Convergence.}
\STATE \# Adapt $\pi^S_\theta$ to target domain; 
\FOR{patient $p$ in $P$}
\FOR{$t = 1, ..., T$} 
\STATE Compute the optimal action $a_t^S=\pi_\theta^S(s_t)$ on source domain;
\STATE Compute the predicted next state $f^S(s_t, a_t^S)$ on source domain;
\FOR{action $a$ $\leftarrow$ $A$} 
\STATE Compute the state distance $||f^T(s_t, a) - f^S(s_t, a_t^S)||$; 
\ENDFOR
\STATE Recommend the action with minimal state distance;
\ENDFOR 
\ENDFOR 
\end{algorithmic} 
\end{algorithm}

\section{Experiments}

To evaluate the performance of the proposed model, we conduct comprehensive comparison experiments on three datasets, including two real-world EHR datasets and a semi-synthetic dataset.


\subsection{Datasets}
\textbf{Real-world datasets.}
Both MIMIC-III\footnote{\url{https://mimic.physionet.org/}} and AmsterdamUMCdb\footnote{\url{https://amsterdammedicaldatascience.nl}} are publicly available real-world EHR datasets. 
Following \cite{mechnical_ventilation},  we extract all adult patients undergoing invasive ventilation more than 24 hours and extract a set of 48 variables, including demographics, vital signs and laboratory values. Following \cite{mechnical_ventilation}, 
We learn the DTR policies for positive end-expiratory pressure (PEEP), fraction of inspired oxygen (FiO2) and ideal body weight-adjusted tidal volume (Vt). We discretize the action space into $7\times 7 \times 7$ actions, as Table \ref{tab:dose_range} shown. 
The statistics of extracted data from MIMIC-III and AmsterdamUMCdb are displayed in Table \ref{tab:stati}.  More details of data preprocessing (e.g., the list of extracted variables) can be found in GitHub\textsuperscript{\ref{github}}.

\begin{table} [!t]
\centering
\caption{Statistics of MIMIC-III and AmsterdamUMCdb} 
\label{tab:stati} 
\begin{tabular}{ccc}
\toprule
 & MIMIC & AmsterdamUMCdb \\
 \midrule
 \#. of patients & 10,843 & 6,560 \\
 \#. of male & 5,931 & 3,412\\
 \#. of female & 4,912 & 3,148\\
 Age (mean $\pm$ std) & 60.7 $\pm$ 11.6 & 62.1 $\pm$ 12.3 \\
 Mortality rate & 24\% & 35\% \\
 \bottomrule
\end{tabular} 
\end{table}

\begin{table*} [!t]
\centering
\caption{Construction of the action space.} 
\label{tab:dose_range} 
\begin{tabular}{cccccccc}
\toprule
  & 1 & 2 & 3 & 4 & 5 & 6 & 7\\
 \midrule
 Vt (mL/Kg) & 0–2.5 & 2.5–5 & 5–7.5 & 7.5–10 & 10–12.5 & 12.5–15 & >15 \\
 PEEP (cmH2O) &  0–5  & 5–7 &  7–9 &  9–11 &  11–13 &  13–15 &  >15 \\
 FiO2 (\%)  &  25–30  & 30–35  & 35–40 &  40–45 &  45–50 &  50–55 &  >55\\
\bottomrule
\end{tabular} 
\vspace{-4pt}
\end{table*}

\vspace{5pt}
\noindent
\textbf{Semi-synthetic dataset based on MIMIC-III.} As the MIMIC-III dataset is real-world observational data, it is impossible to obtain the potential outcomes for underlying counterfactual treatment actions. To evaluate the proposed model's ability to learn optimal DTR policies, we further validate the method in a simulated environment. 
We simulate hidden state $s_t$ and observational data $o_t$ for each patient at time $t$ following $p$-order autoregressive process \cite{mills1991time}. The details of the simulation can be found in supplementary material and GitHub\textsuperscript{\ref{github}}.   

\begin{table*} [h]
\centering
\caption{Performance comparison for policy evaluation on test sets. 
Note that RL and CI denote reinforcement learning and causal inference respectively.} 
\label{tab:res}  
\begin{tabular}{llcc|cc|cc}
\toprule  
 &  & \multicolumn{2}{c}{MIMIC} 
 & \multicolumn{2}{c}{AmsterdamUMCdb} 
 & \multicolumn{2}{c}{Semi-synthetic} \\ 
 \cline{3-8}
 &  & EM $\downarrow$ & WIS $\uparrow$ & ~~EM~~ $\downarrow$ &  ~~WIS~~ $\uparrow$ & ACC-3$\uparrow$ & ACC-1$\uparrow$ \\ 
 \hline  
  & Imitation Learning $^S$            & 0.21 & 1.85  & 0.26 & 1.21  & 0.31 & 0.63\\
Supervised
  & Imitation Learning $^M$            & 0.23 & 1.84  & 0.28 & 0.95  &  0.27 & 0.61\\
 learning
  & Imitation Learning $^A$            & 0.21 & 1.98  & 0.25 & 1.26  & 0.32 & 0.65\\
 
  & MDP                                & 0.22 & 2.04  & 0.26 & 1.03  & 0.28 & 0.62\\
 \hline
 
 \multirow{4}{*}{RL} 
  & AI Clinician \cite{aiclinician}    & 0.19 & 2.15 & 0.24 & 1.45  & 0.34 & 0.68\\
  & VentAI \cite{mechnical_ventilation}    & 0.19 & 2.21 & 0.24 & 1.46  & 0.34 & 0.69\\
  & DQN \cite{dqn}                     & 0.20 & 2.33 & 0.25 & 1.43  & 0.36 & 0.70\\
  & MoE \cite{amia2018}                & 0.19 & 2.29 & 0.24 & 1.40  & 0.36 & 0.69\\
  & SRL-RNN \cite{kdd2018}             & 0.19 & 2.47 & 0.25 & 1.58  & 0.37 & 0.70\\
 \hline
  
 \multirow{2}{*}{RL with CI} 
  & CIQ \cite{ciq}                     & 0.18 & 2.68 & 0.24 & 1.68 & 0.41 & 0.72\\
  
  & CIRL \cite{cirl}                   & 0.18 & 2.70 & 0.23 & 1.65 & 0.42 & 0.73\\
 \hline
 
 \multirow{5}{*}{Ours} 
  & DAC$^{-rsp}$           & 0.18 & 2.75 & 0.23 & 1.78 & 0.42 & 0.74\\
  & DAC$^{-dcf}$      & 0.17 & 2.78 & 0.23 & 1.82 & 0.41 & 0.72\\
  
  & DAC$^{-short}$                     & 0.17 & 2.93 & \textbf{0.22} & 1.89 & 0.44 & 0.74\\
  & DAC$^{-long}$                      & 0.18 & 2.80 & 0.24 &  1.79 & 0.42 & 0.72 \\
  & DAC                               & \textbf{0.16} & \textbf{3.13} & \textbf{0.22} & \textbf{2.03} & \textbf{0.45} & \textbf{0.76}\\ 

\bottomrule 
\end{tabular} 
\end{table*}

\subsection{Methods for comparison} 
We compare the proposed model with following baselines.

\noindent
\textbf{Supervised models:}
\begin{itemize}
    \item \textbf{Markov Decision Process (MDP)}: The observations of variables are clustered into 750 discrete
    mutually exclusive patient states with k-means. Markov decision process is used to learn the state transition matrix with different actions.  Only the discharged patients are used during the training phase. 
    
    \item \textbf{Imitation Learning}: Imitation learning models the patient states with LSTM, and mimics the human clinician policy. Different from MDP, the hidden states of LSTM can represent continuous states of patients. We implemented three versions of imitation learning by training the same model on different datasets. Imitation Learning~$^S$ is trained on the discharged patients. Imitation Learning~$^M$ is trained on the patients with 90-day mortality. Imitation Learning~$^A$ is trained on all the patients in the training set.
\end{itemize}

\noindent
\textbf{RL-based DTR learning models:}
\begin{itemize}
    \item \textbf{AI Clinician} \cite{aiclinician}: AI clinician clustered patient states into 750 groups and adopts MDP to model the patient state transition. The difference between AI clinician and MDP is that AI clinician model is trained based on Q-learning while MDP only mimics the human clinician strategy. 

    \item \textbf{VentAI} \cite{mechnical_ventilation}: VentAI also adopts MDP to model the patient state transition and uses Q-learning to learn optimal policies for mechanical ventilation.

    \item \textbf{DQN} \cite{dqn}: DQN leverages LSTM to model patient health states, and Q-learning is used to train the dynamic treatment regime learning model. 
    
    \item \textbf{Mixture-of-Experts (MoE)} \cite{amia2018}: MoE is a mixture model of a neighbor-based policy learning expert (kernel) and a model-free policy learning expert (DQN). The mixture model switches between kernel and DQN experts depending on patient's current history.
    
    \item \textbf{SRL-RNN} \cite{kdd2018}: SRL-RNN is based on actor-critic framework. LSTM is used to map patients' temporal EHRs into vector sequences. The model combines the indicator signal and evaluation signal through joint supervised and reinforcement learning. 
\end{itemize}

\noindent
\textbf{RL-based models with causal inference:}
\begin{itemize}
    \item \textbf{Causal inference Q-network (CIQ)} \cite{ciq}: CIQ trains Q-network with interfered states and labels.
    Gassian noise and adversarial observations are considered in the training of CIQ.
    
    \item \textbf{Counterfactual inverse reinforcement learning (CIRL)} \cite{cirl}: 
    CIRL learns to estimate counterfactuals and integrates counterfactual reasoning into batch inverse reinforcement learning.  
    
\end{itemize}

\noindent
\textbf{Variants of DAC:}
We implement the proposed model with five versions.
DAC is the main version. By removing the patient resampling module, confounding balance module, long-term rewards or short-term rewards, we train another four versions DAC$^{-rsp}$, DAC$^{-dcf}$, DAC$^{-long}$, DAC$^{-short}$ to conduct the ablation study.

Note that the extracted variables contain lots of vital signs and lab values, which have lots of missing values. The baselines can only take fixed-sized observed variables as inputs. Following \cite{aiclinician,raghu2019reinforcement}, we impute the missing values with multi-variable nearest-neighbor imputation~\cite{imputation} before training the baseline models.

\begin{figure*}
\centering 
\subfigure[]{
\includegraphics[width=0.28\textwidth]{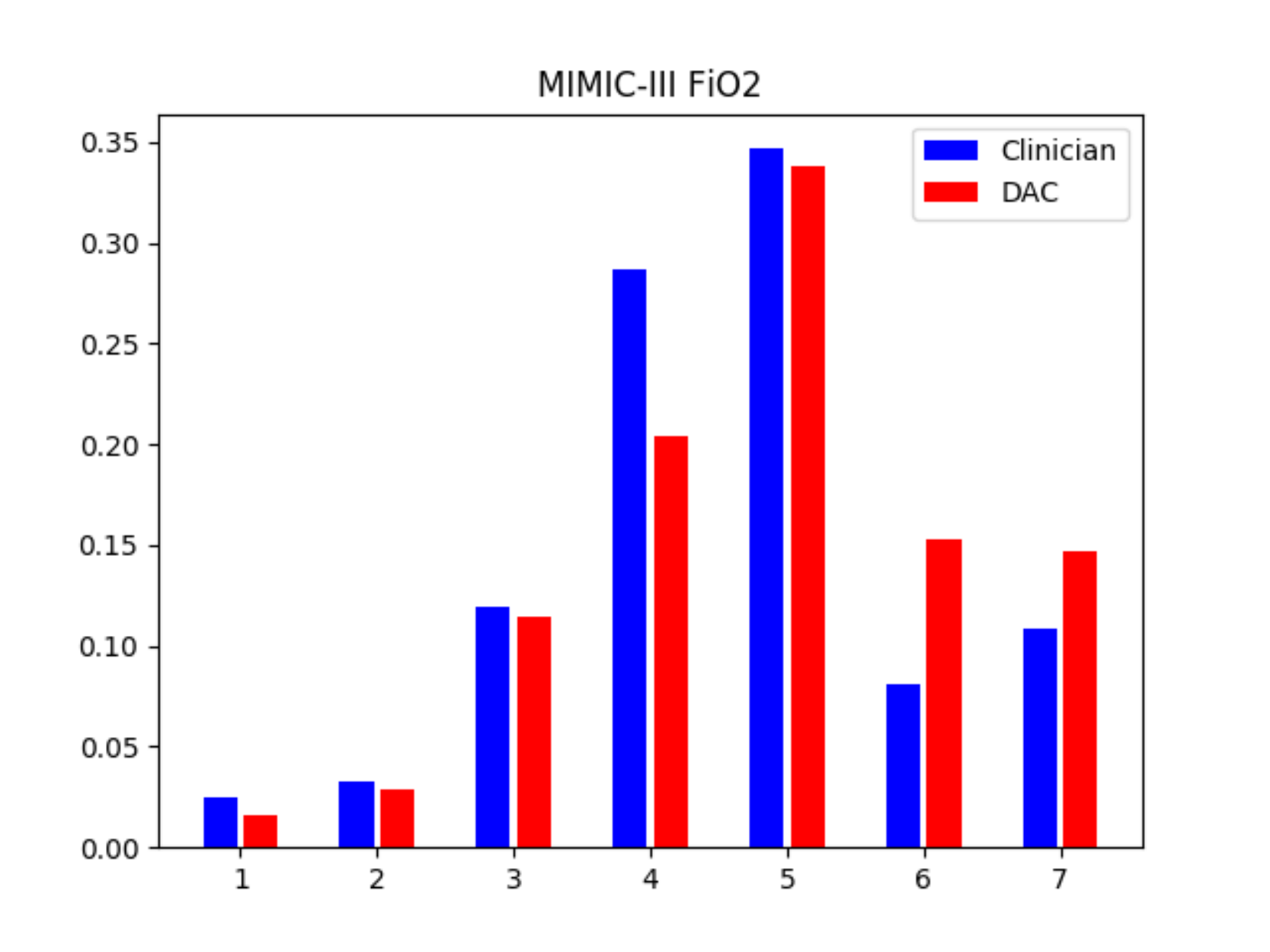}} 
\subfigure[]{
\includegraphics[width=0.28\textwidth]{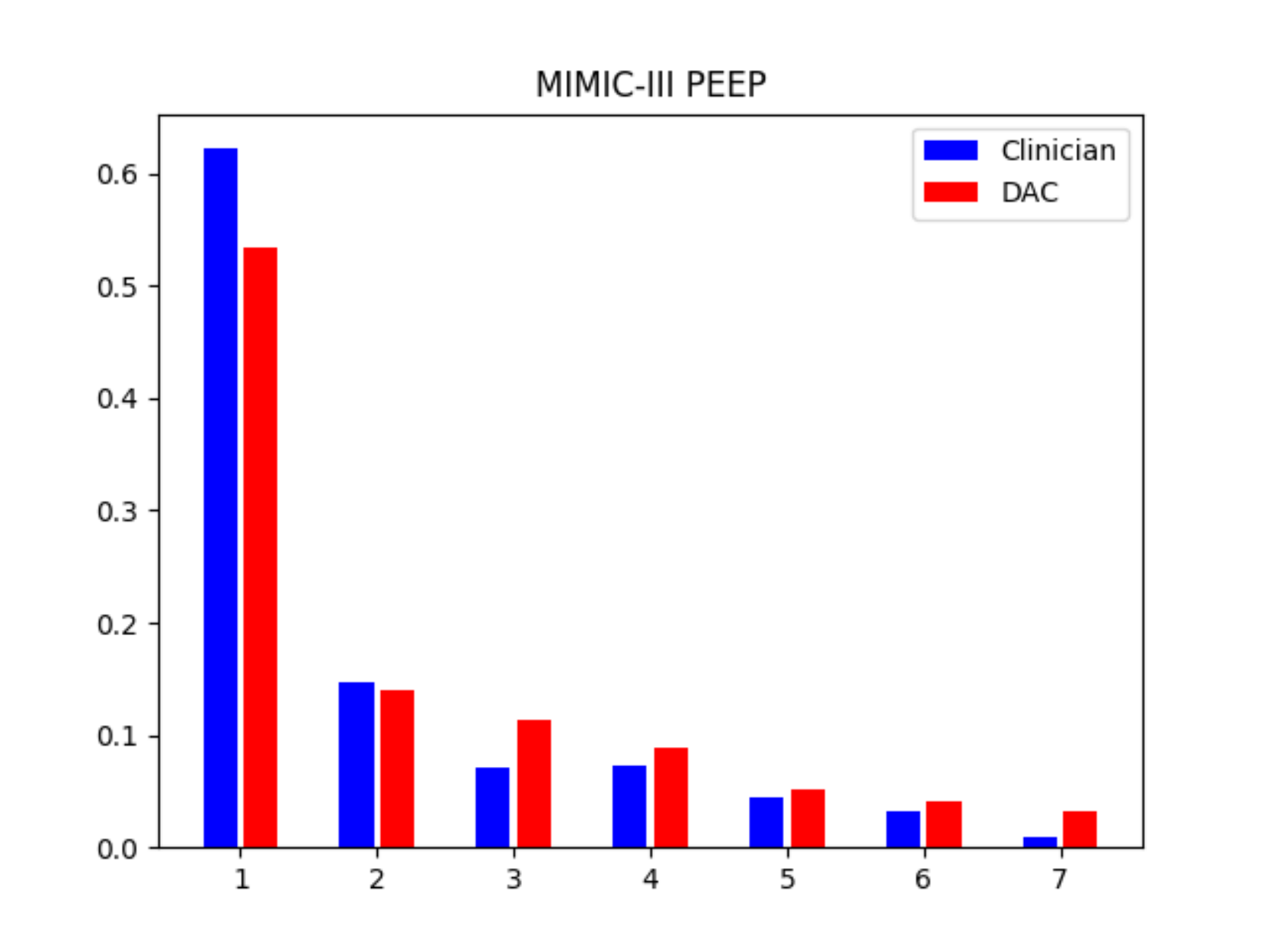}}
\subfigure[]{
\includegraphics[width=0.28\textwidth]{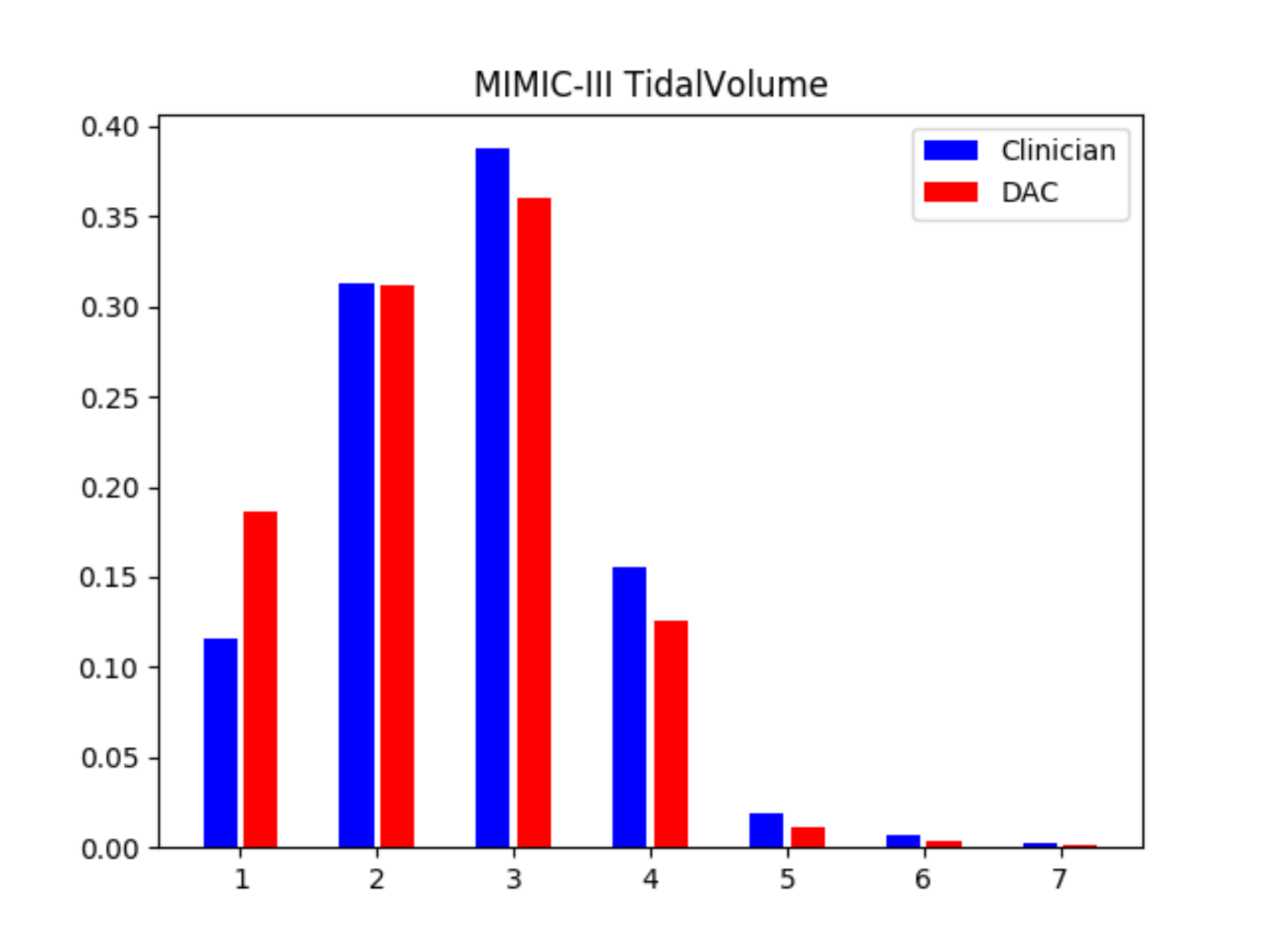}} 
\caption{Visualization of the action distribution in the 3-dimensional action space on MIMIC-III dataset.  The horizontal axis denotes the discritized actions and the vertical axis denotes the distribution of corresponding actions.} 
\label{fig:distribution_mimic} 
\end{figure*}

\noindent
\textbf{Implementation details.} 
We implement our proposed model with Python 2.7.15 and PyTorch 1.3.0. For training models, we use Adam optimizer with a mini-batch of 256 patients. The observed variables and corresponding values are projected into a $512$-d space. The models are trained on 1 GPU (TITAN RTX 6000), with a learning rate of 0.0001. We randomly divide the datasets into 10 sets. All the experiment results are averaged from 10-fold cross validation, in which 7 sets were used for training every time, 1 set for validation and 2 sets for test. The validation sets are used to determine the best values of parameters in the training iterations. 
More details and implementation code are available in GitHub\textsuperscript{\ref{github}}.

Note that there are $7 \times 7 \times 7=343$ kinds of actions three parameters (i.e., PEEP, FiO2 and tidal volume). At the beginning of training phase, it might be inaccurate to compute the probabilities of 343 kinds of actions, which would cause the computed weight in Eq.~(\ref{eq:weight}) to be unstable. 
Moreover, clinical guidelines \cite{guideline_fan2017official,guideline_lahouti} also recommend clinicians to increase or decrease the parameters according to patients' health states. When computing the inverse probabilities, we use the probabilities for 3 action changes (i.e., increase, decrease or keep the same for each parameter) instead of the probabilities of 7 actions.

\subsection{Evaluation Metrics}
\noindent
\textbf{Evaluation metrics.}
The evaluation metrics for treatment recommendation in real-world datasets is still a challenge \cite{kdd2018,rl_evaluation}. Following  \cite{kdd2018,raghu2017deep,weng2017representation,aiclinician,zhang2017leap}, we try two evaluation metrics estimated mortality (\textbf{EM}), weighted importance sampling (\textbf{WIS}) to compare the proposed model with the state-of-art methods for real-world datasets. In the simulated environment, we have access to the ground truth of optimal actions and compute the optimal action accuracy rate following \cite{cirl,ciq}. Mechanical ventilator has three important parameters: PEEP, Vt and FiO2. We compute two kinds of accuracy rates: \textbf{ACC-3} (whether the three parameters are set the same as the optimal action simultaneously) and \textbf{ACC-1} (whether each parameter is set correctly).  The details of the metric calculation can be found in supplementary material.

\subsection{Result Analysis}

Table \ref{tab:res} displays the estimated mortality, WIS and action accuracy rates on the three datasets. The results show that the proposed model outperforms the baselines. 
The RL-based models (e.g., AI Clinician, MoE, SRL-RNN) achieve lower estimated mortality rates and higher WIS and action accuracy rates than supervised models (i.e., Imitation Learning and MDP), which demonstrates the effectiveness of RL in DTR learning tasks.

Among the three versions of imitation learning,  Imitation Learning $^M$ is trained on the non-survivor patients and thus performs worse than the other two versions. However, Imitation Learning $^M$ still achieves comparable performance to MDP trained on discharged patients, which demonstrates the clinicians' treatment strategies for survivors and non-survivors are similar. Thus it is not appropriate to directly punish the treatment actions prescribed to patients with mortality. We speculate that for some non-survivors, the treatment actions might be helpful but the confounder (e.g., the poor health states before treatments) caused the 90-day mortality. Thus we propose deconfounding modules to alleviate the patient state distribution bias.
Taking into account the confounders, CIQ, CIRL and the proposed models outperform the RL baselines, which demonstrates the effectiveness of incorporation of counterfactual reasoning in DTR learning tasks. Among the models with the consideration of confounders, the proposed DAC performs better than CIQ and CIRL. We speculate the reasons are two-fold: (i) we train DAC on balanced mini-batch by resampling the patients, which makes critic network's counterfactual action reward estimation more accurate; (ii) the proposed short-term rewards are more efficient at capturing short-term patients' health state changes than discounted long-term rewards during the training of RL models.

Among the five versions of the proposed model, the main version (i.e., DAC) outperforms DAC$^{-rsp}$ and DAC$^{-dcf}$, which demonstrates the effectiveness of proposed patient resampling and confounding balance modules. Combining short-term and long-term rewards, DAC outperforms DAC$^{-short}$ and DAC$^{-long}$, which demonstrates the effectiveness of the two designed rewards.

\vspace{5pt}
\noindent
\textbf{Distribution of Actions}: Visualization of the action distribution in the 3-D action space on MIMIC-III are shown in Figure \ref{fig:distribution_mimic}. The results show that our model learned similar policies to clinicians on MIMIC-III dataset. DAC suggests more actions with the higher PEEP and FiO2. Besides, the learning policies recommend more frequent lower tidal volume compared to clinician policy.

\begin{figure*}
\centering 
\subfigure[]{
\includegraphics[width=0.28\textwidth]{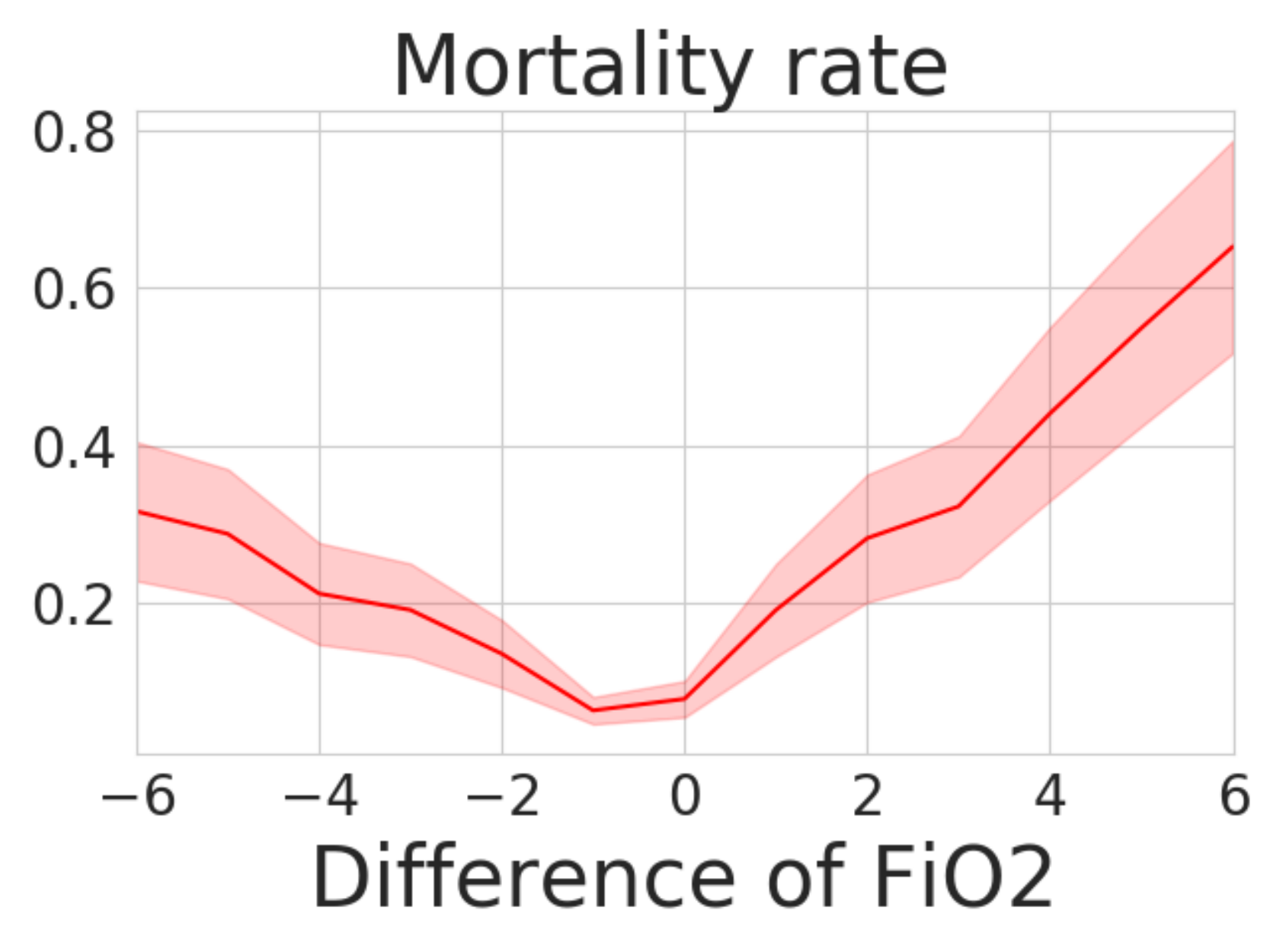}} 
\subfigure[]{
\includegraphics[width=0.28\textwidth]{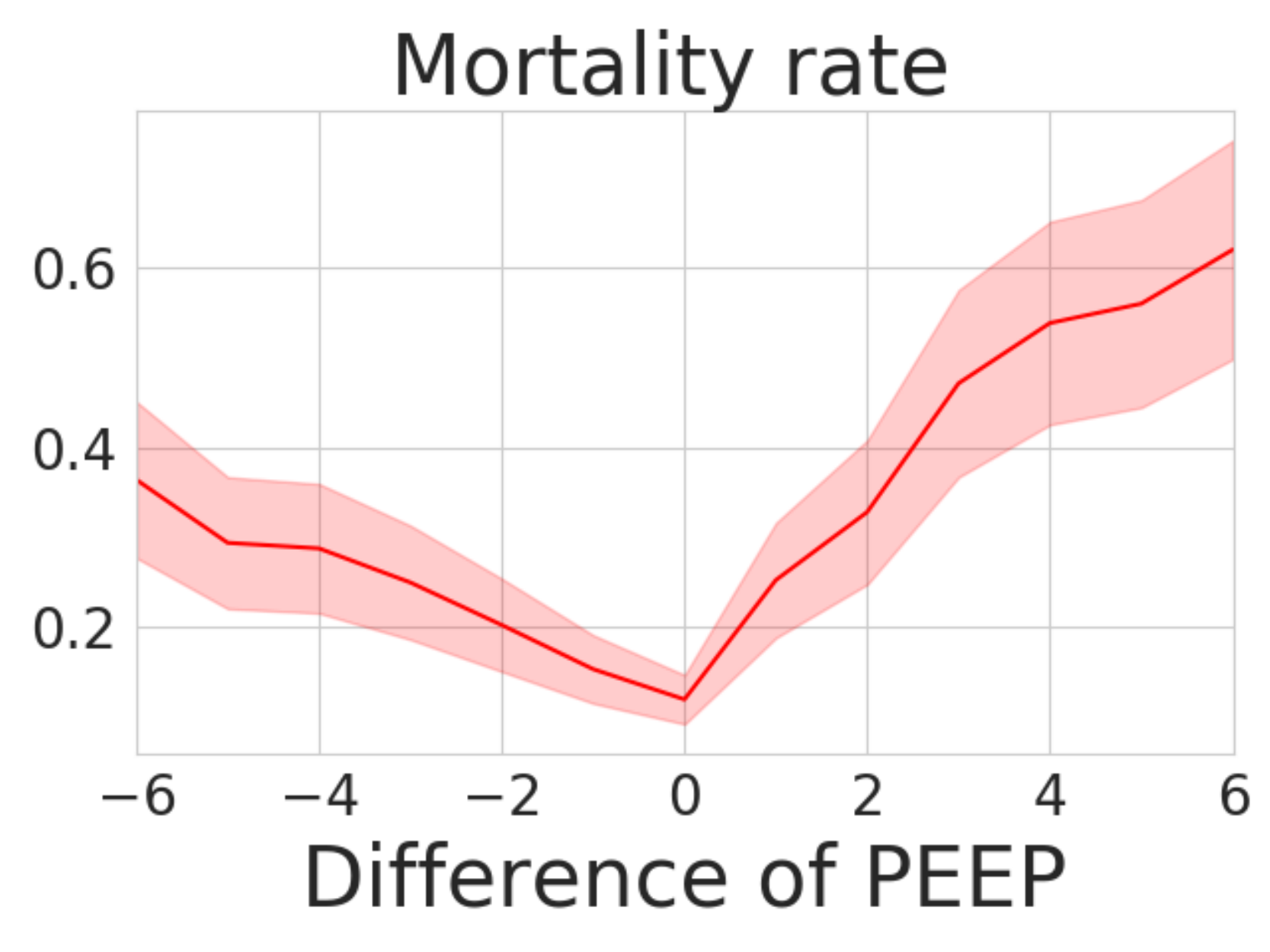}}
\subfigure[]{
\includegraphics[width=0.28\textwidth]{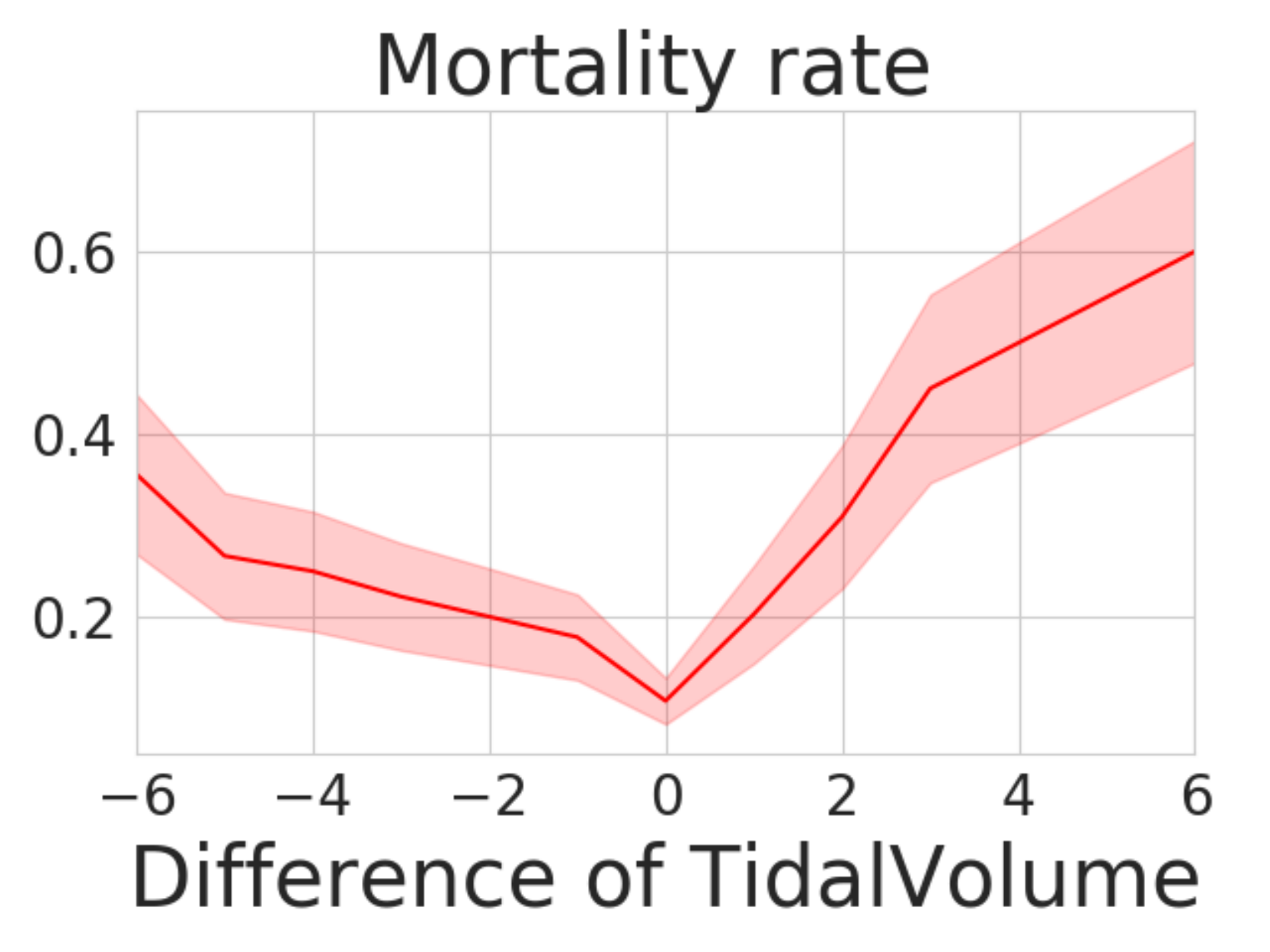}} 
\caption{The relations between mortality rates and mechanical ventilation setting difference (recommended setting - actual setting) on MIMIC-III dataset. } 
\label{fig:dose_diff_mimic} 
\end{figure*}

\noindent
\textbf{Comparison of Clinician and DAC policies}: 
We find that the mortality rates are lowest in patients for whom clinicians’ actual treatments matched the actions the learned policies recommend. Figure~\ref{fig:dose_diff_mimic} shows the relations between mortality rate and mechanical ventilation setting difference on MIMIC-III. The results show when patients received lower values of FiO2, PEEP and Tidal Volume, the mortality rates increase much faster. We speculate the reasons are two-fold: (i) DAC only recommends high values of FiO2, PEEP and Tidal volume to the high-risk patients, who still have relatively higher mortality rates even with optimal treatments; (ii) the high-risk patients received low-value settings, which further increased their mortality rates.

\begin{figure}
 \centering 
\subfigure[]{
\includegraphics[width=0.23\textwidth]{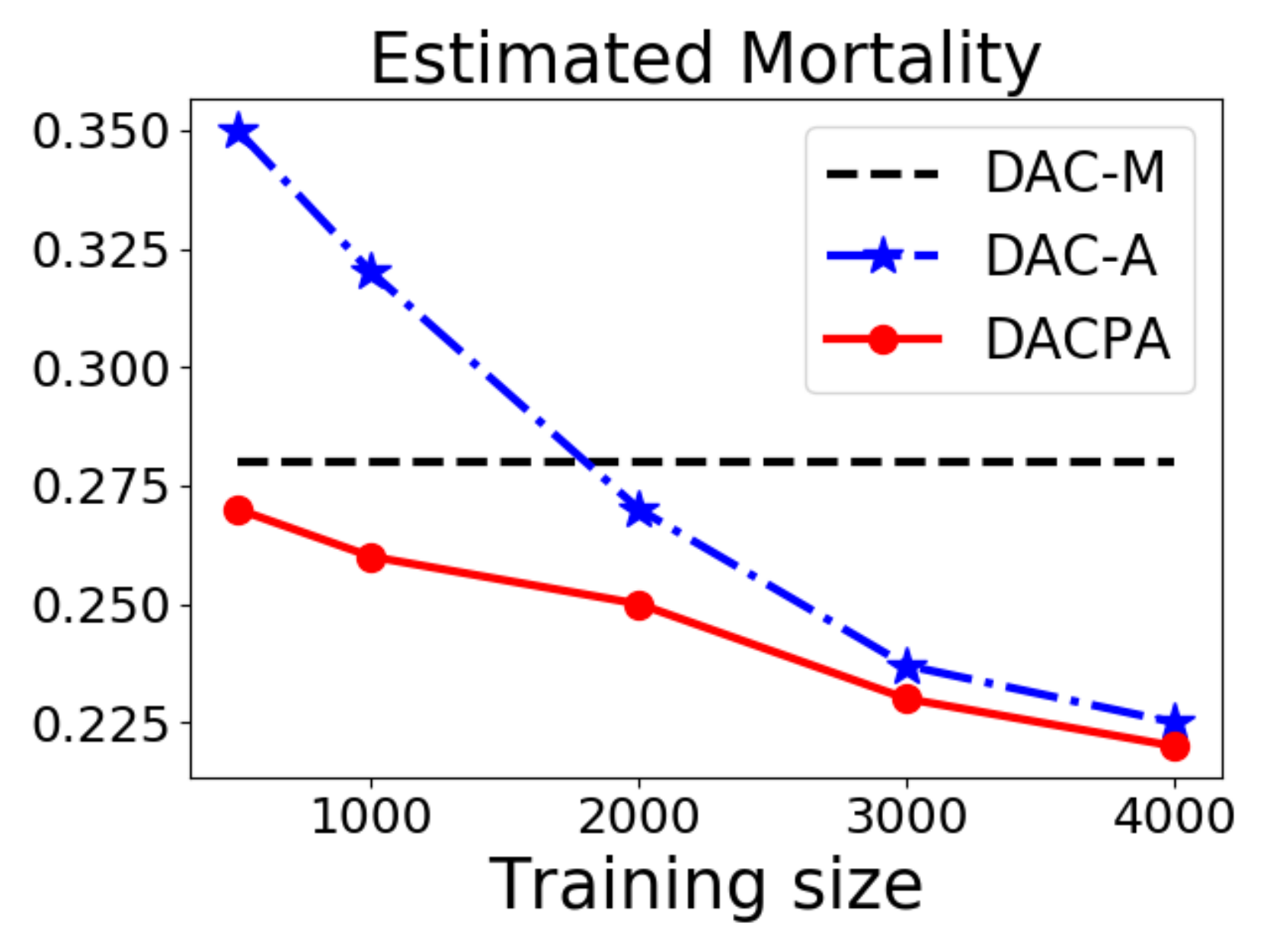}}
\subfigure[]{
\includegraphics[width=0.23\textwidth]{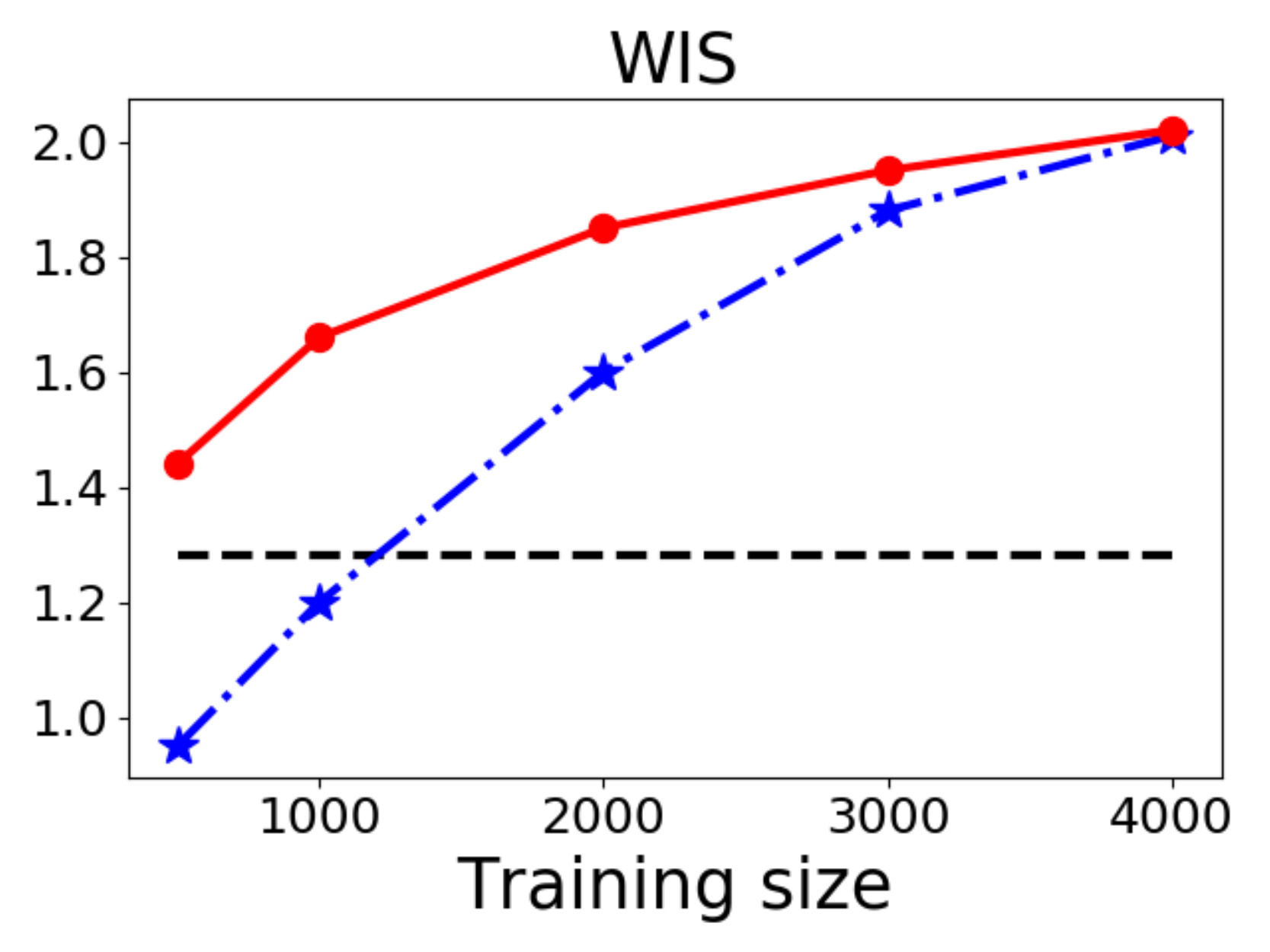}} 
\caption{Performance of policy adaptation to AmsterdamUMCdb dataset over different training size. } 
\vspace{-4pt}
\label{fig:transfer} 
\end{figure}

\vspace{5pt}
\noindent
\textbf{Policy adaptation}: We adapt the model trained on MIMIC-III to AmsterdamUMCdb, and Fig. \ref{fig:transfer} shows the estimated mortality and WIS with different training sizes on AmsterdamUMCdb. DAC-M is trained on MIMIC-III and directly validated on AmsterdamUMCdb. DAC-A is trained on AmsterdamUMCdb and DACPA is pretrained on MIMIC-III and then adapted to AmsterdamUMCdb. The results show that with transfer learning on AmsterdamUMCdb, DACPA outperforms DAC-M, which demonstrates that the policy adaption is very helpful and improved model performance. When training size becomes smaller, the performance gaps between DACPA and DAC-A are larger, which demonstrates that the introduced policy adaption method is useful when adapting trained models to new-source small-scale datasets.

\section{Related Work}
In this section, we briefly review the existing works related to DTR and causal inference.  

\vspace{5pt}
\noindent
\textbf{DTR learning.}
During recent years, there have been some studies that focus on applying RL to the optimal treatment learning from existing (sub)optimal clinical datasets. 
Komorowski et al. \cite{aiclinician} proposed AI Clinician model which uses a Markov decision process (MDP) to model patients' health states and learns the treatment strategy to prescribe vasopressors and IV fluids with Q-learning. 
Raghu et al. \cite{mechnical_ventilation} uses a similar model to AI Clinician to learn the optimal DTR policies for mechanical ventilation and achieves lower estimated mortality rates than human clinicians.
\cite{raghu2017deep} expands on Komorowski’s initial work by proposing a Dueling Double Deep Q network Q-learning model with a continuous state space and introduces a continuous reward function to train the model. They show that for patients with higher severity of illness, due to a lack of data, the model did not outperform the human clinicians.
\cite{amia2018} presents mixture-of-experts (MoE) to combine the restricted DRL approach with a kernel RL approach selectively based on the context and find that the combination of the two methods achieves a lower estimated mortality rate. \cite{kdd2018} proposes a new Supervised Reinforcement Learning with Recurrent Neural Network (SRL-RNN) model for dynamic treatment regime, which combines the indicator signal and evaluation signal through the joint supervised learning and RL. The experiments demonstrate that the introduced supervised learning is helpful for stably learning the optimal policy. 
Although the DTR learning algorithms can achieve high performance on treatment recommendation tasks, the learned policies could be biased without the consideration of confounding issues. 

\vspace{5pt}
\noindent
\textbf{DTR learning with causal inference.}
Causal inference \cite{greenland1999causal,pearl2009causality,robins1995analysis} has been used to empower the learning process under noisy observation and can provide interpretability for decision-making models \cite{schulam2017reliable,bica2020estimating,atan2018deep,bica2020time,johansson2016learning}. In this paper, we focus on the related work of DTR learning with causal inference. Zhang and Schaar \cite{zhang2020gradient} propose a gradient regularized V-learning method to learn the value function of DTR with the consideration of time-varying confounders. Bica et al. \cite{bica2020estimating} present a  Counterfactual Recurrent Network (CRN) to estimate treatment effects over time and recommend optimal treatments to patients.
Yang et al. \cite{ciq} investigates the resilience ability of an RL agent to withstand adversarial and potentially catastrophic interferences and proposed a causal inference Q-network (CIQ) by training RL with additional inference labels to achieve high performance in the presence of interference. Bica et al. \cite{cirl} propose a counterfactual inverse reinforcement learning (CIRL) by integrating counterfactual reasoning into batch inverse reinforcement learning. From a conceptual point of view, the studies most closely related to ours are \cite{ciq,cirl} and we compare the proposed DAC with them. Both two studies incorporate causal inference into RL models. The key difference between ours and theirs are: (i) we resample the patients and the training DAC with balanced mini-batch can improve the model performance; (ii) we design a short-term reward that can further remove the confounding;
(iii) our model is based on actor-critic framework and the critic network can provide more accurate rewards with the help of the patient resampling module and short-term reward;
(iv) we introduce a policy adaptation method to the proposed DAC, which can efficiently adapt trained models to new-source environments.

\section{Conclusion}

In this paper, we investigate the confounding issues and data imbalance problem in clinical settings, which could limit optimal DTR learning performance of RL models. The training of most existing DTR learning methods is guided by the long-term clinical outcomes (e.g., 90 day mortality), so some optimal treatment actions in the history of non-survivors might be punished. 
To address the issues, we propose a new deconfounding actor-critic network (DAC) for mechanical ventilation dynamic treatment regime learning.
We propose a patient resampling module and a confounding balance module to alleviate the confounding issues.  
Moreover, we introduce a policy adaptation method to the proposed DAC to transfer the learned DTR policies to new-source datasets. Experiments on a semi-synthetic   dataset and two publicly available real-world datasets (i.e., MIMIC-III and AmsterdamUMCdb) show that the proposed model outperforms state-of-the-art methods, demonstrating the effectiveness of the proposed framework.  
The proposed model can provide individualized treatment decisions that could improve patient outcomes.

\bibliographystyle{ACM-Reference-Format}
\bibliography{main}

\appendix

\clearpage

\section{Semi-synthetic dataset based on MIMIC-III}

As the MIMIC-III dataset is real-world observational data, it is impossible to obtain the potential outcomes for underlying counterfactual treatment actions. To evaluate the proposed model's ability to learn optimal DTR policies, we further validate the method in a simulated environment. 
The treatment assignments $a_t$ at each time stamp are influenced by the confounders $q_t$, which are consist of hidden confounders $s_{t}$ and time-varying covariates $o_t$. We first simulate $o_t$ and $s_{t}$ for each patient at time $t$ following $p$-order autoregressive process \cite{mills1991time} as,
\begin{equation}
\begin{aligned}
    o_{t,j} = \frac{1}{p}\sum_{r=1}^{p}(\alpha_{r,j}o_{t-r,j}+\beta_{r}a_{t-r}) + \eta_{t}\\
    s_{t,j} = \frac{1}{p}\sum_{r=1}^{p}(\mu_{r,j}s_{t-r,j}+\upsilon_{r}a_{t-r}) + \epsilon_{t}
\end{aligned}
\end{equation}
where $o_{t,j}$ and $s_{t,j}$ denote the $j$-th column of $o_t$ and $s_{t}$, respectively. For each $j$, $\alpha_{r,j},\mu_{r,j}\sim \mathcal{N}(1-(r/p),(1/p)^{2})$ control the amount of historical information of last p time stamps incorporated to the current representations. $\beta_{r},\upsilon_{r}\sim \mathcal{N}(0, 0.02^{2})$ controls the influence of previous treatment assignments. $\eta_{t},\epsilon_{t}\sim \mathcal{N}(0,0.01^{2})$ are randomly sampled noises. 

To simulate the treatment assignments, we generate $10,000$ survivor patients and $30,000$ non-survivor patients. 
The confounders $q_t$ at time stamp $t$ and outcome $y$ can be simulated using the hidden confounders and current covariates as follows,
\begin{equation}
\label{eq:synthetic-outcome}
\begin{aligned}
& q_t =  \frac{1}{t}\sum_{r=1}^{t}s_{r} +  g(o_t)\\
& y = w^{\top}q_{T}+b
\end{aligned}
\end{equation} 
where   $w\sim\mathcal{U}(-1,1)$ and $b\sim\mathcal{N}(0,0.1)$. The function $g(\cdot)$ maps  $o_t$ into the hidden space. 

\section{Evaluation Metrics}
The evaluation metrics for treatment recommendation is still a challenge \cite{kdd2018,rl_evaluation}. Thus we try different evaluation metrics to compare the proposed model with the state-of-art methods. 

\begin{figure}[h]
\centering
\subfigure[]{
\includegraphics[width=0.23\textwidth]{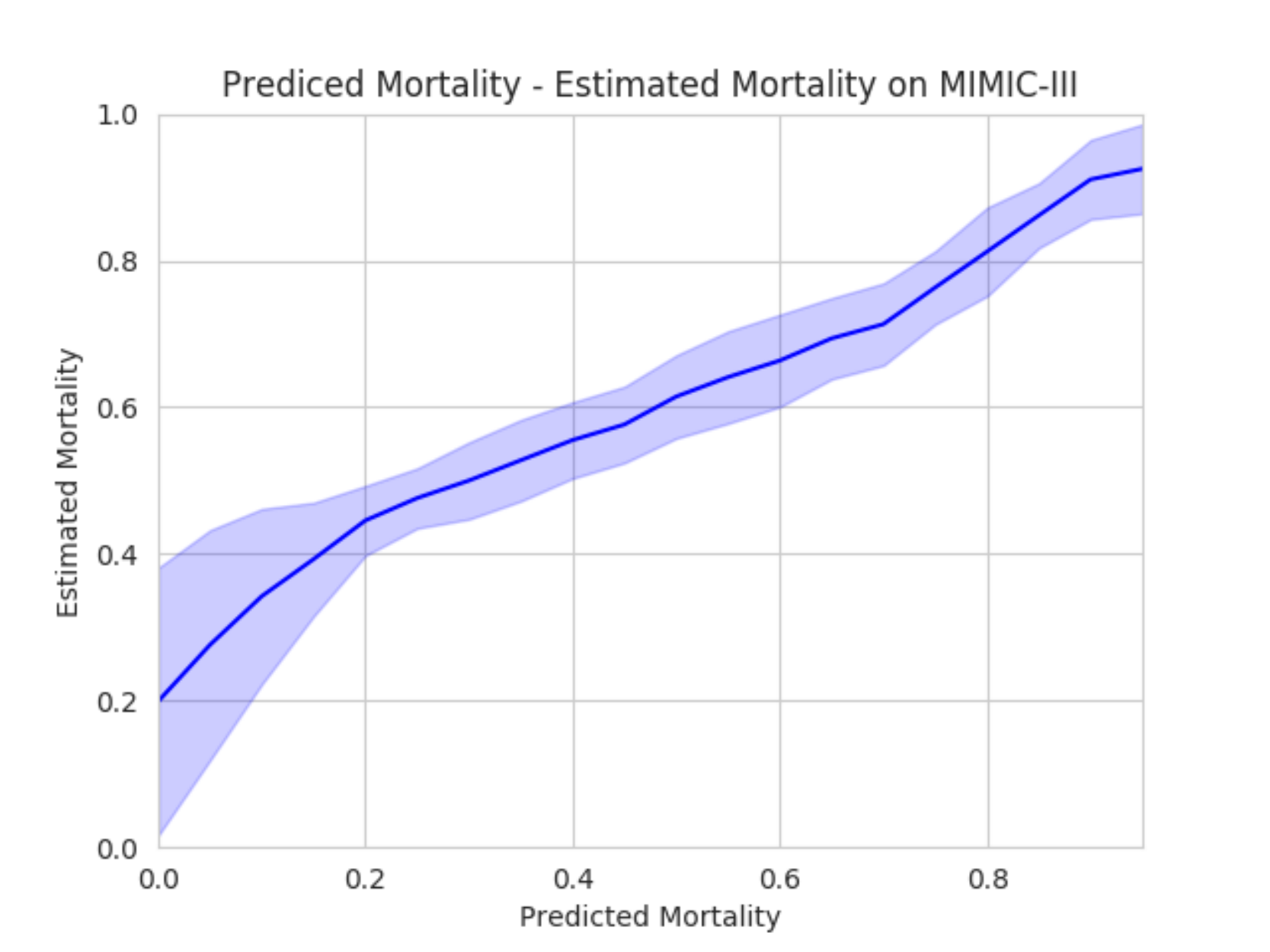}}
\subfigure[]{
\includegraphics[width=0.23\textwidth]{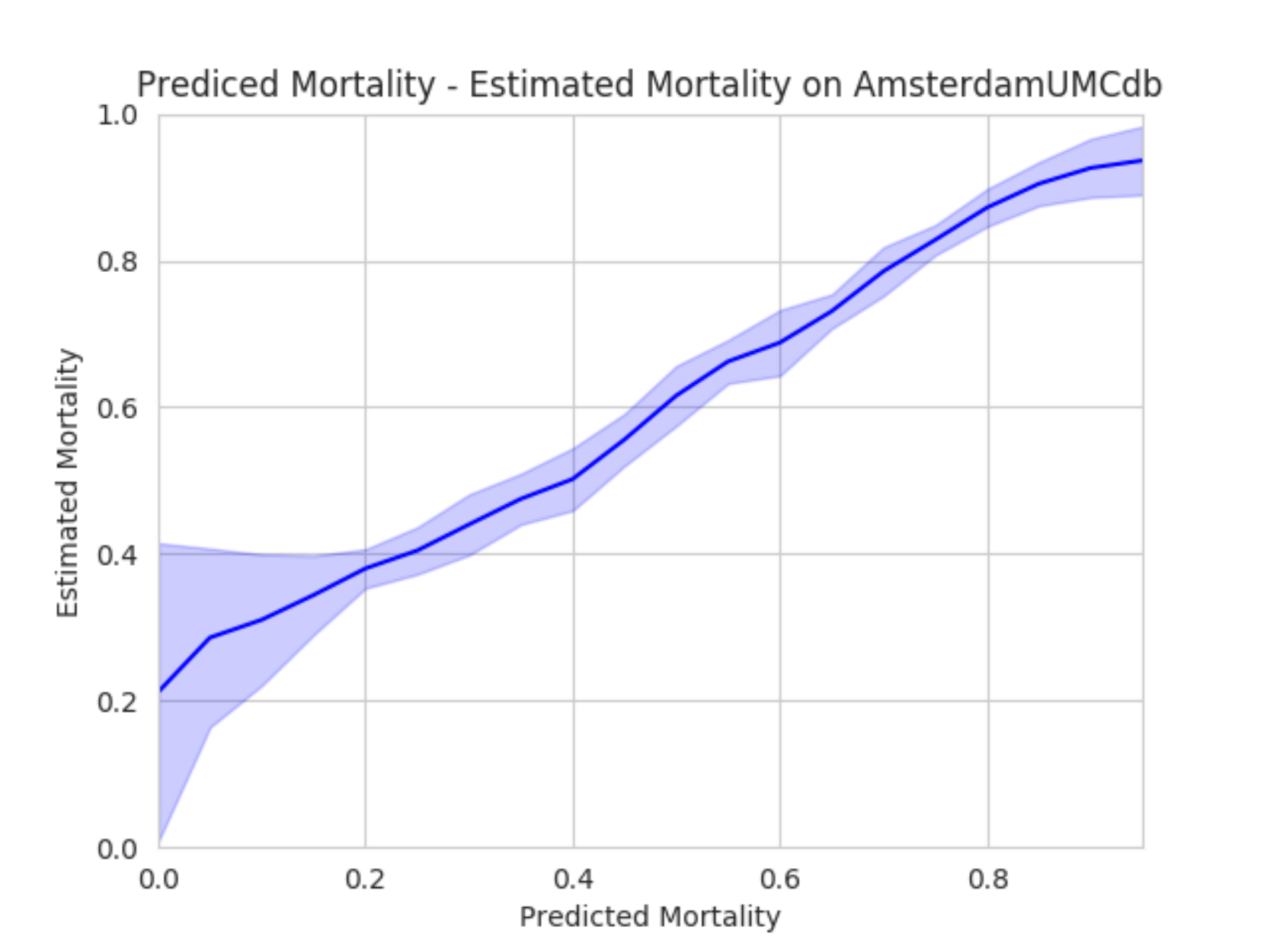}}
\vspace{-4mm}
\caption{The positive correlations between estimated mortality rate and predicted mortality probability on MIMIC-III and AmsterdamUMCdb datasets. } 
\label{fig:predicted_mortality}
\end{figure}

\noindent
\textbf{Estimated mortality}: Following \cite{kdd2018,raghu2017deep,weng2017representation}, we use the estimated in-hospital mortality rates to measure whether policies would eventually reduce the patient mortality or not. 
Specifically, we train a mortality risk prediction model, which takes the patient states and next actions as inputs, and output mortality risks. The predicted mortality risks are discretized into different units with small intervals shown in the x-axis of Figure \ref{fig:predicted_mortality}. Discharged patients dominate both datasets, so the predicted mortality rates are smaller than the actual mortality rate in the real-world clinical setting. We adjusted the predicted mortality rate to calculate a new estimated mortality rate. 
Given an example denoting an admission of a patient, if the patient died in hospital, all the predicted mortality rates belonging to this admission are associated with a value of mortality and the corresponding units add up these values. After scanning all test examples, the average estimated mortality rates for each unit are calculated, shown in y-axis of Figure \ref{fig:predicted_mortality}. Based on these
results, the estimated mortality rates corresponding to the predicted mortality rate of different policies are used as the measurements to denote the estimated in-hospital mortality. Although the estimated mortality does not equal the mortality in the real-world clinical setting, it is a universal metric currently for computational testing. The relations between estimated mortality rate and predicted mortality probability are shown in Figure \ref{fig:predicted_mortality}.

\noindent
\textbf{Weighted importance sampling(WIS)}: Following \cite{aiclinician,raghu2017deep}, we also implement  a high-confidence off-policy evaluation (HCOPE) method (WIS). The human clinician policy is defined as $\pi_0$, and $\pi_1$ denotes the learned AI policy. We defined $\rho_t = \pi_1(a_t, s_t)/\pi_0(a_t,s_t)$ as the per-step importance ratio, where $(a_t, s_t)$ represent the $t^{th}$ actual (action, state) pair for a patient. $\rho_{1:t} = \pi_{t'=1}^t \rho_{t'}$ is the cumulative importance ratio up to step $t$ and $w_t = \sum_{i=1}^{|D|} \rho_{1:t}^{(i)}/ |D|$ denotes the average cumulative importance ratio at horizon $t$ in dataset $D$ and $|D|$ as the number of trajectories in $D$. The trajectory-wise WIS estimator is given by:
\vspace{-4mm} 
\begin{equation}
\label{eq:vwis} 
V_{WIS} = \frac{\rho_{1:H}}{w_H}(\sum_{t=1}^H \gamma^{t-1}R_t),
\vspace{-1mm}
\end{equation}
where $H$ denotes the length of steps for the patient and $R_t$ denote the long-term reward. 
Then, the WIS estimator is the average estimate over all trajectories, namely:
\vspace{-2mm}
\begin{equation}
\label{eq:wis} 
WIS = \frac{1}{|D|}\sum_{i=1}^{|D|} V_{WIS}^{(i)},
\vspace{-2mm}
\end{equation} 
where $V_{WIS}^{(i)}$ is WIS applied to the trajectory for $i^{th}$ patient.

\begin{figure*}[tbp]
\centering
\subfigure[]{
\includegraphics[width=0.25\textwidth]{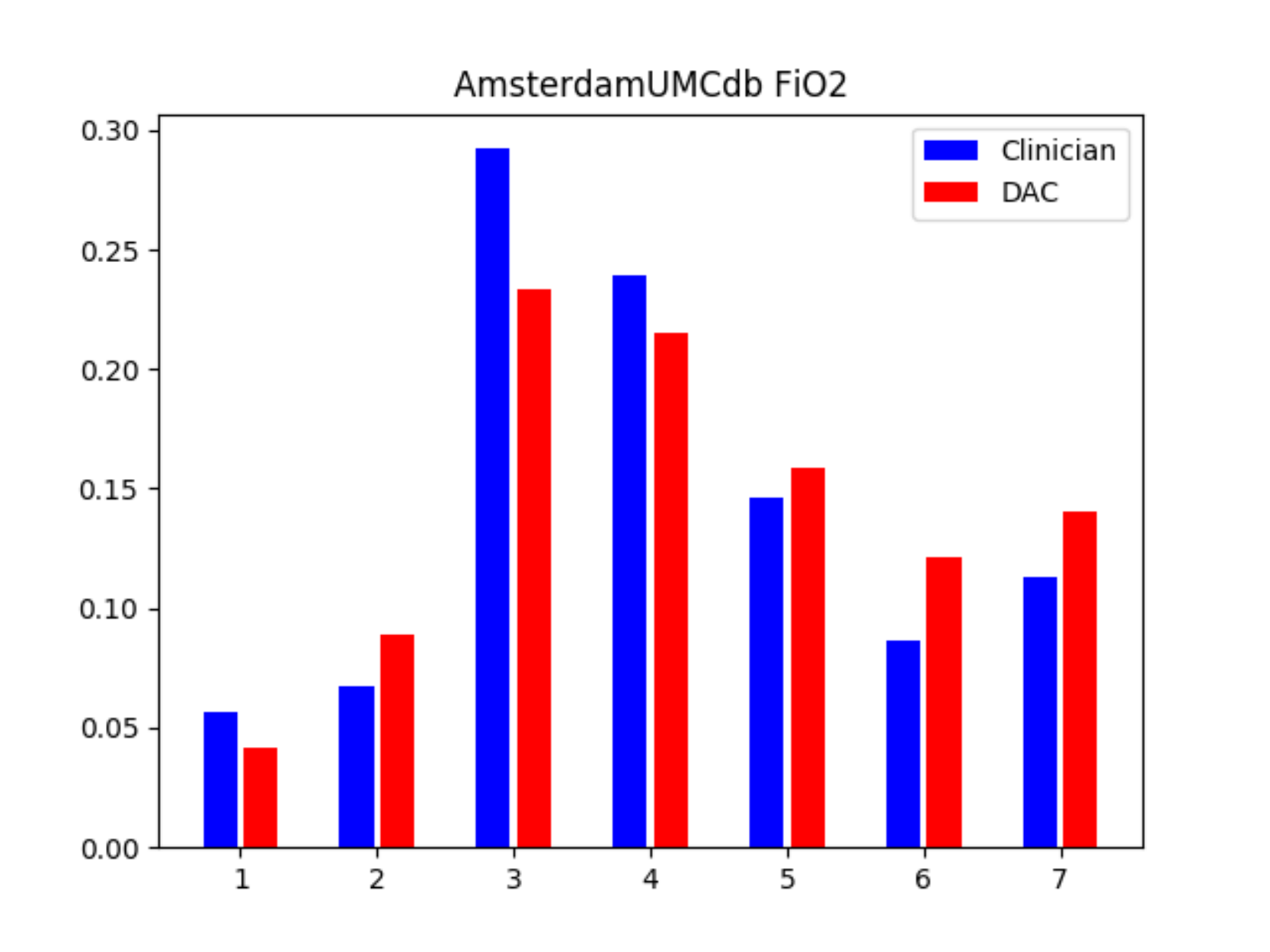}}
\subfigure[]{
\includegraphics[width=0.25\textwidth]{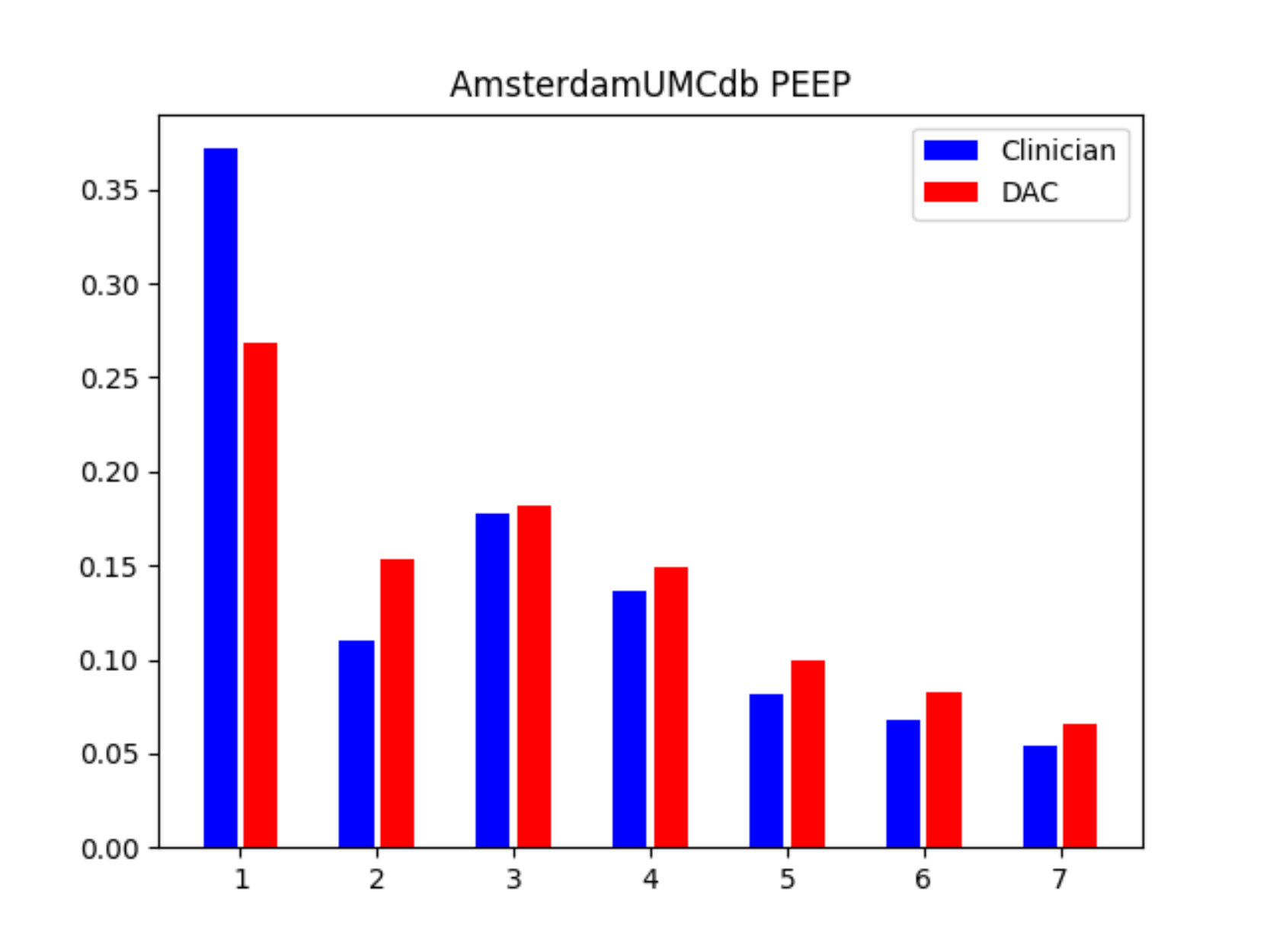}}
\subfigure[]{
\includegraphics[width=0.25\textwidth]{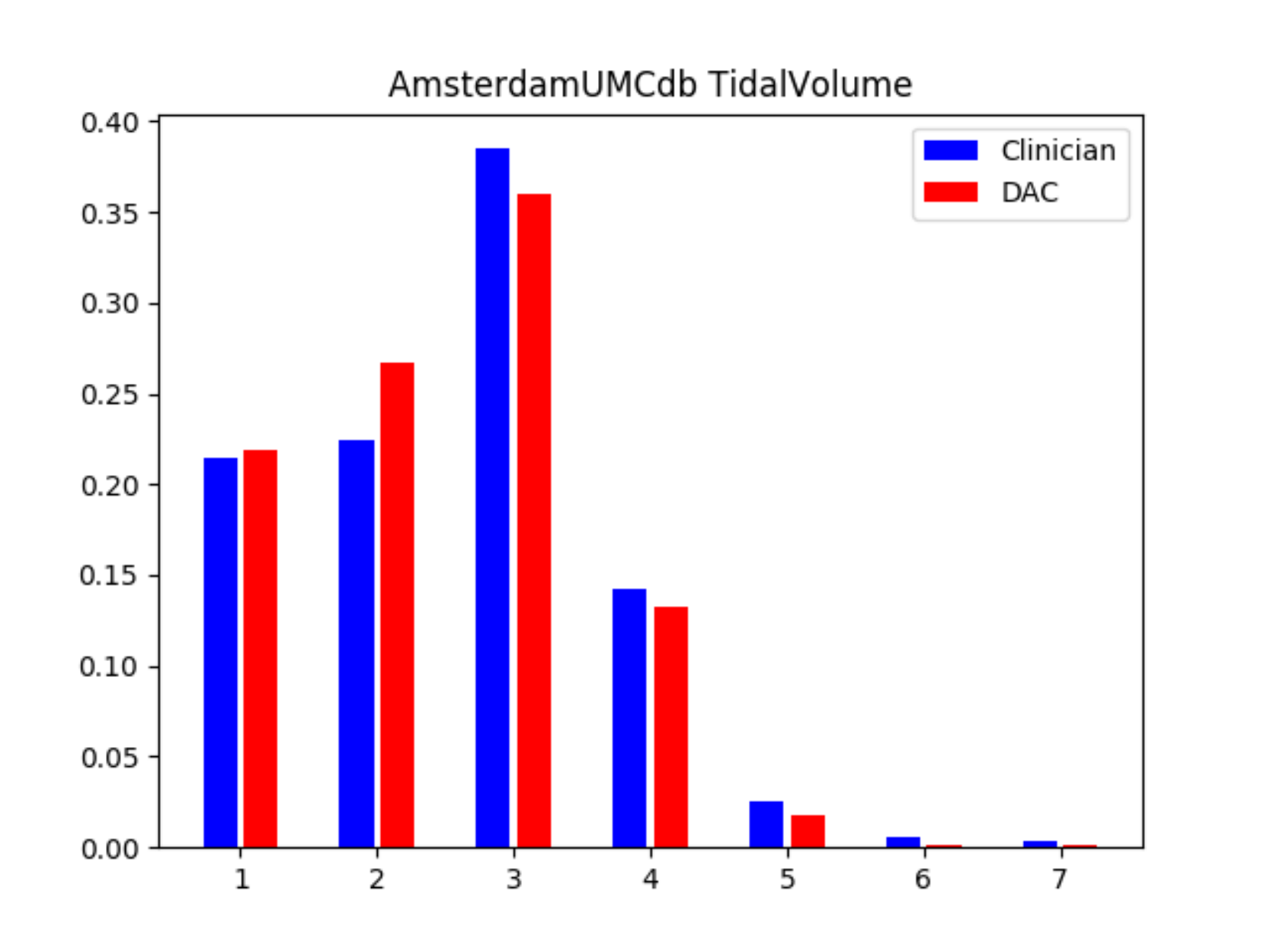}} 
\vspace{-10pt}
\caption{Visualization of the action distribution in the 3-dimensional action space on AmsterdamUMCdb.} 
\label{fig:distribution_ast}
\end{figure*}

\begin{figure*}[!ht]
\centering
\subfigure[]{
\includegraphics[width=0.25\textwidth]{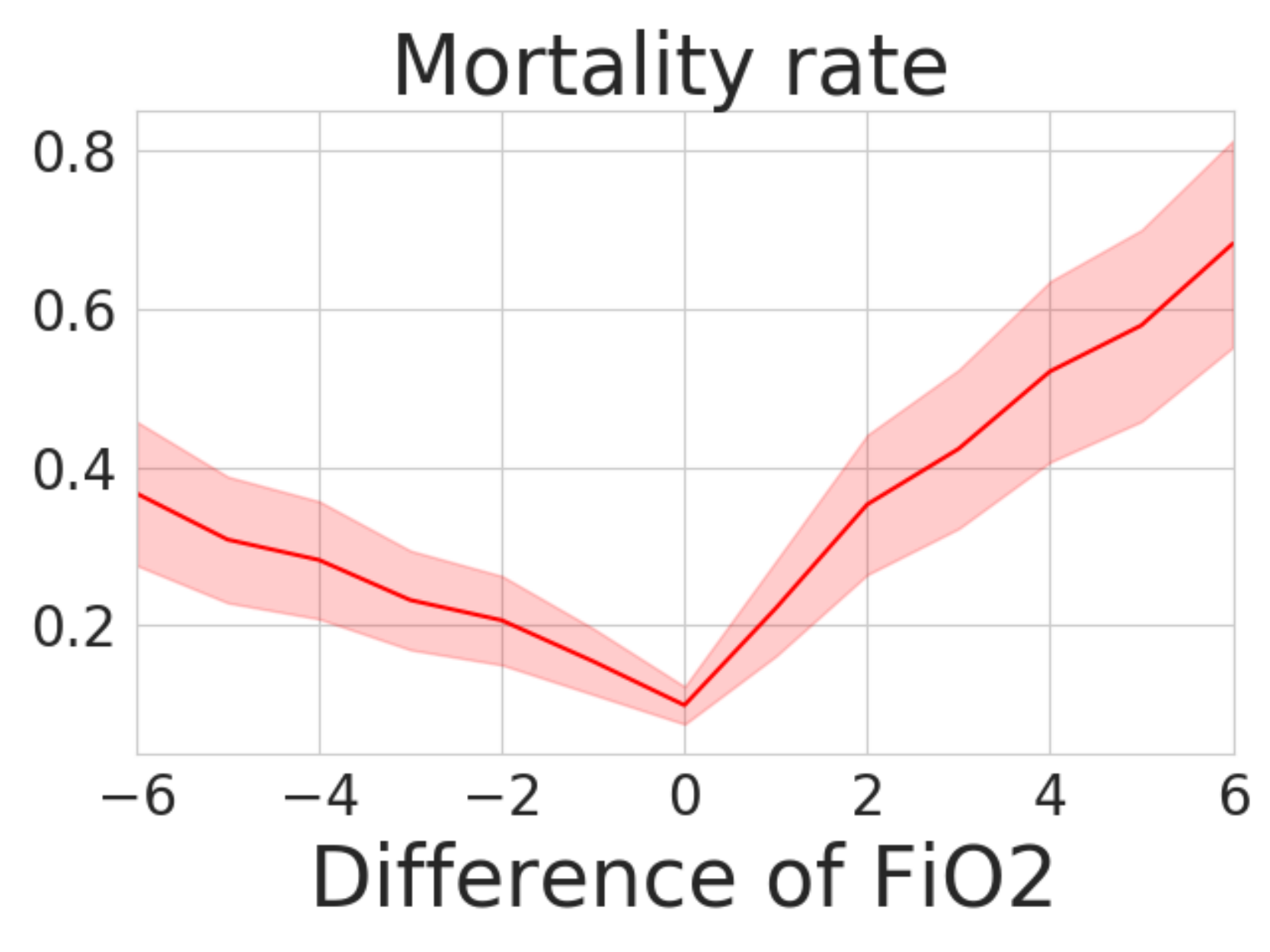}} 
\subfigure[]{
\includegraphics[width=0.25\textwidth]{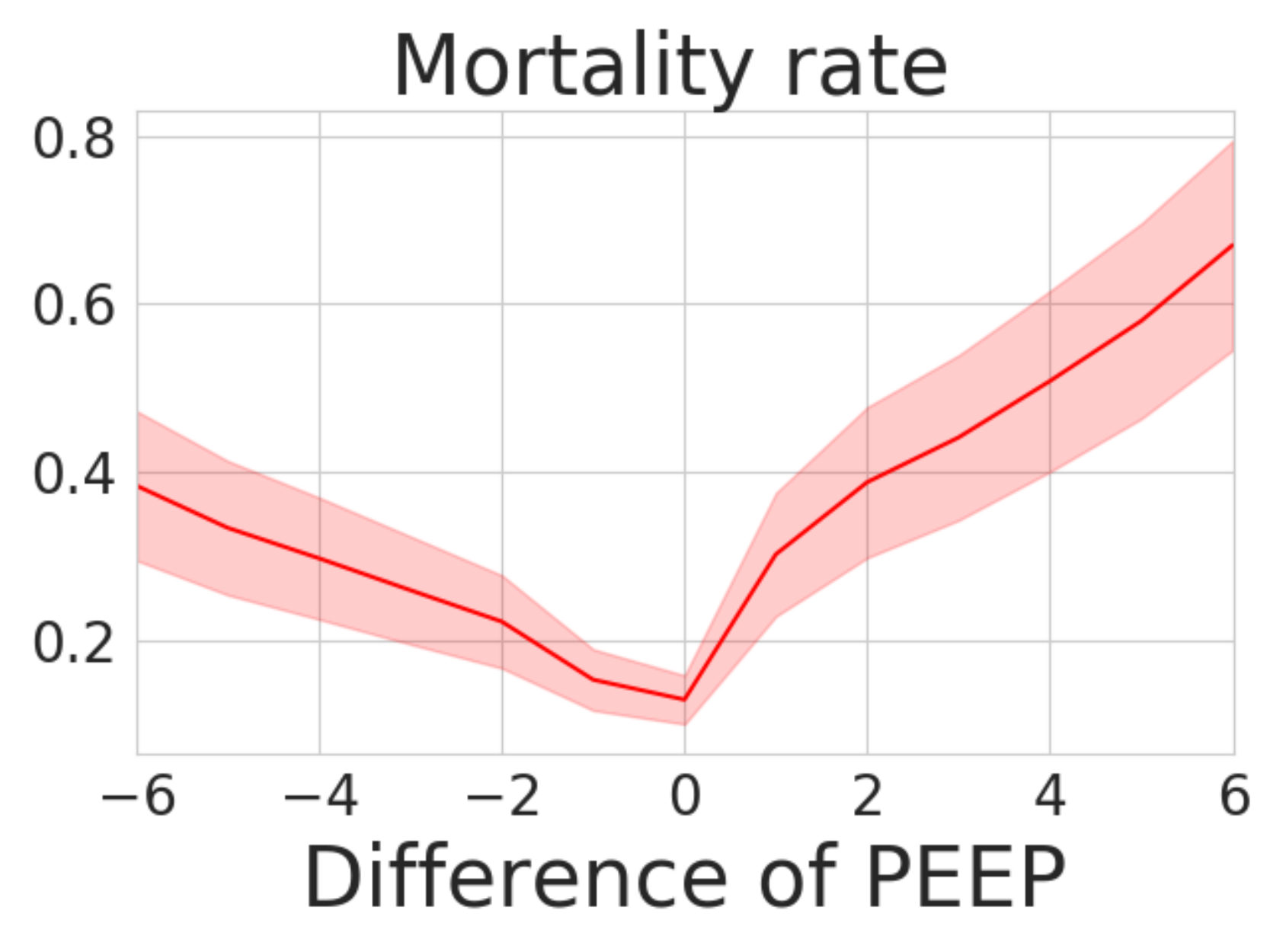}}
\subfigure[]{
\includegraphics[width=0.25\textwidth]{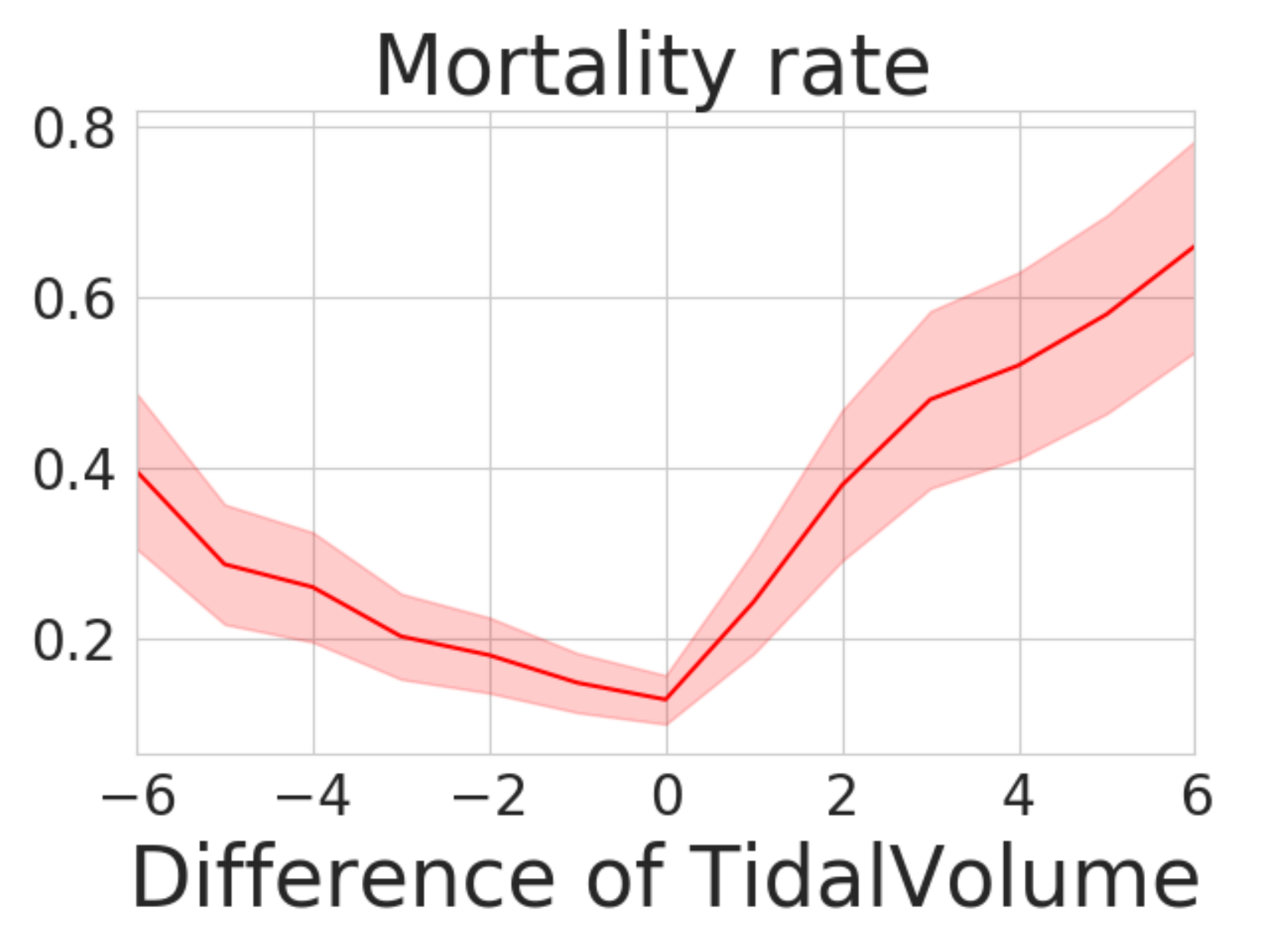}}
\vspace{-10pt}
\caption{The relations between mortality rate and medicine dose gaps between human clinician and DAC policies on AmsterdamUMCdb. } 
\label{fig:dose_diff_ast}
\end{figure*}

\begin{figure}[ht]
\centering
\subfigure[]{
\includegraphics[width=0.23\textwidth]{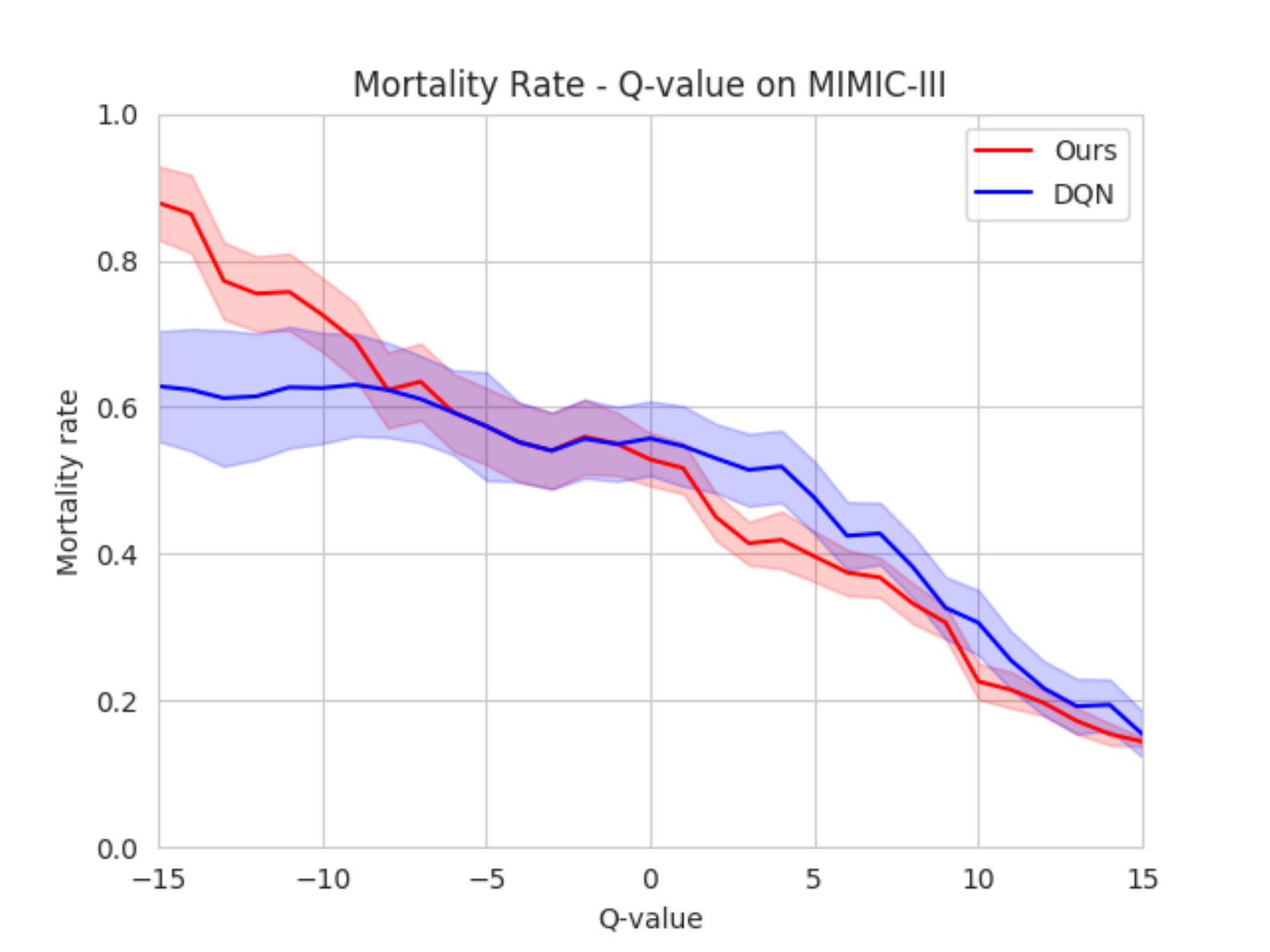}}
\subfigure[]{
\includegraphics[width=0.23\textwidth]{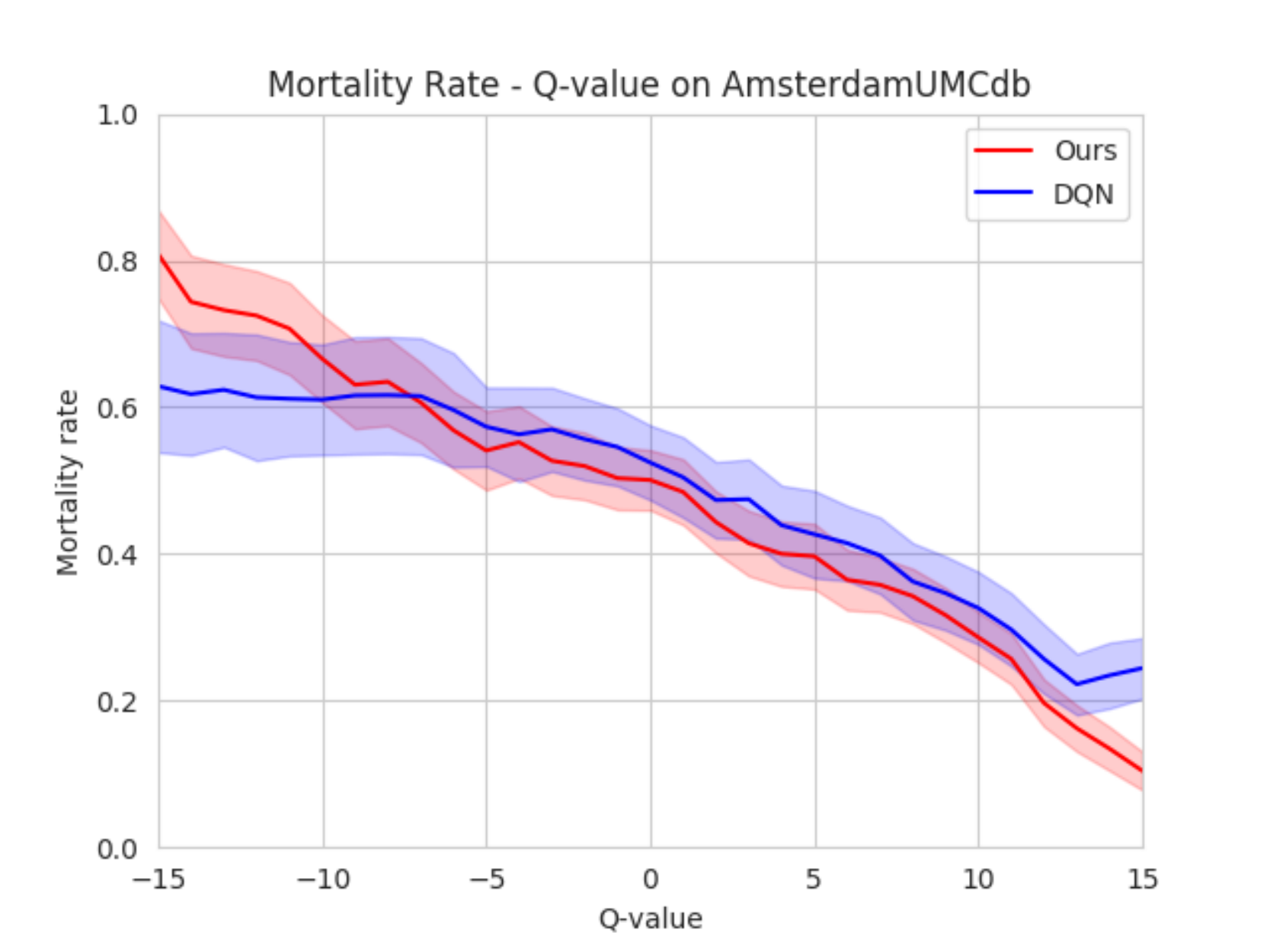}}
\vspace{-4mm}
\caption{ Mortality-expected-return curve computed by the learned policies} 
\label{fig:q_value}
\end{figure}

\noindent
\textbf{Action accuracy rate}: 
Following \cite{cirl,ciq}, we compute the optimal action accuracy rate to evaluate the models' performance to learn optimal DTR policies in simulated environments.
Mechanical ventilator has three important parameters: PEEP, Vt and FiO2. We compute two kinds of accuracy rates: \textbf{ACC-3} (whether the three parameters are set the same as the optimal action simultaneously) and \textbf{ACC-1} (whether each parameter is set correctly).  The metrics are computed as follows:
\begin{equation*}
\label{eq:acc-3} 
ACC-3 = \frac{1}{|D|}\sum_{i=1}^{|D|} \frac{1}{T}\sum_{t=1}^T f(a_t^p, \hat{a}_t^p) * f(a_t^v, \hat{a}_t^v) * f(a_t^f, \hat{a}_t^f),
\end{equation*} 
\begin{equation}
\label{eq:acc-1} 
ACC-1 = \frac{1}{|D|}\sum_{i=1}^{|D|} \frac{1}{T*3}\sum_{t=1}^T f(a_t^p, \hat{a}_t^p) + f(a_t^v, \hat{a}_t^v) + f(a_t^f, \hat{a}_t^f),
\end{equation} 
$$f(a,b)=
\begin{cases}
1& \text{if }a = b\\
0& \text{else }
\end{cases},$$ 
where $a_t^p$, $a_t^v$, $a_t^f$ are recommened actions for PEEP, Vt and FiO2, $\hat{a}_t^p$, $\hat{a}_t^v$, $\hat{a}_t^f$ are optimal actions. 

\section{Additional Experimental Results}

The relations between expected returns and mortality rates are shown in Figure \ref{fig:q_value}. The results show that our model has a more clear negative correlation between expected returns and mortality rates than DQN in both MIMIC-III and AmsterdamUMCdb datasets. The reason might be two-fold: (i) DQN is trained on the initial EHR data with confounder bias; (ii) DQN punishes the actions used for patients who suffer from mortality, while some actions might be optimal.

\begin{figure}[!h]
\centering
\subfigure[]{
\includegraphics[width=0.23\textwidth]{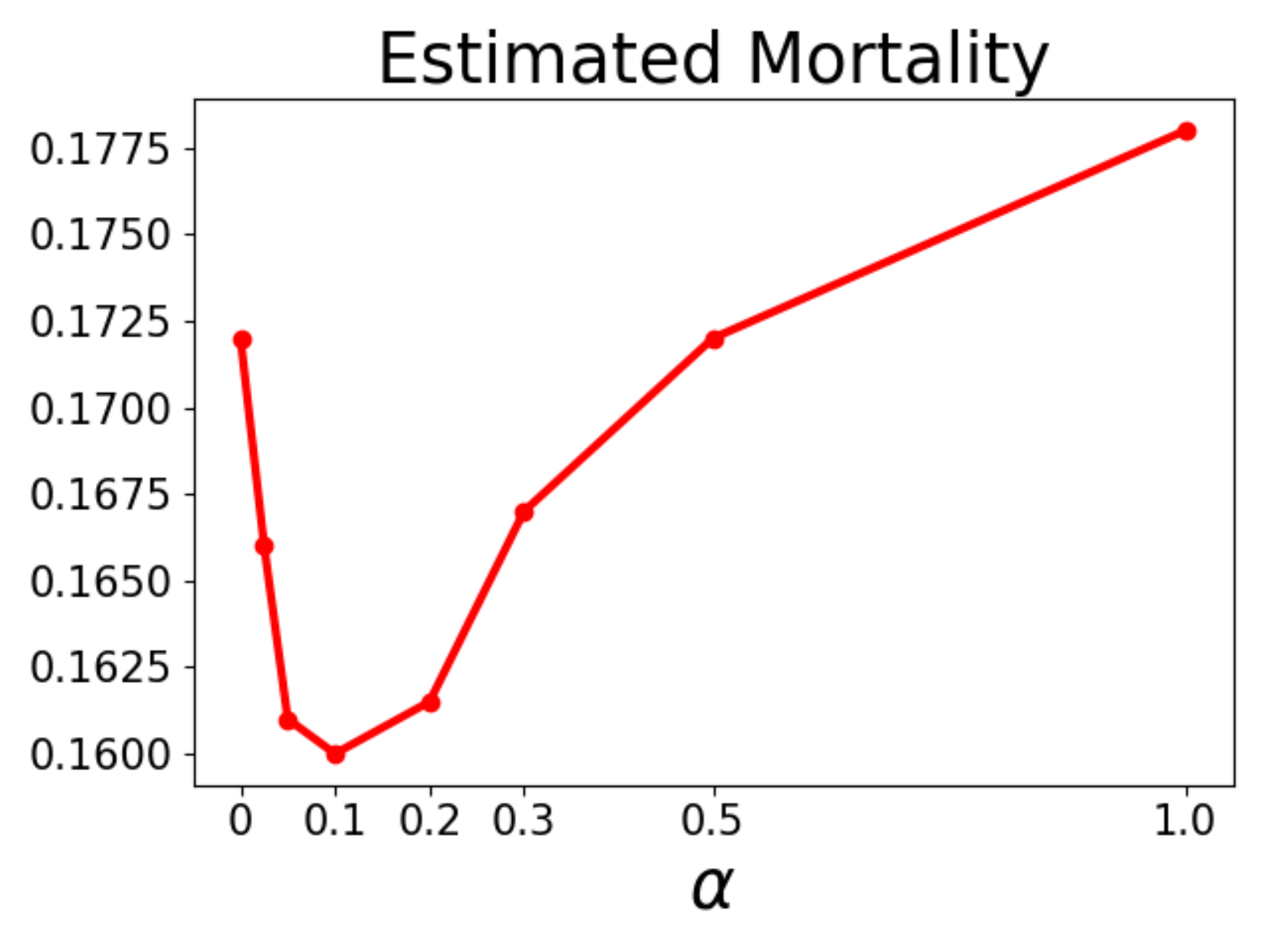}}
\subfigure[]{
\includegraphics[width=0.23\textwidth]{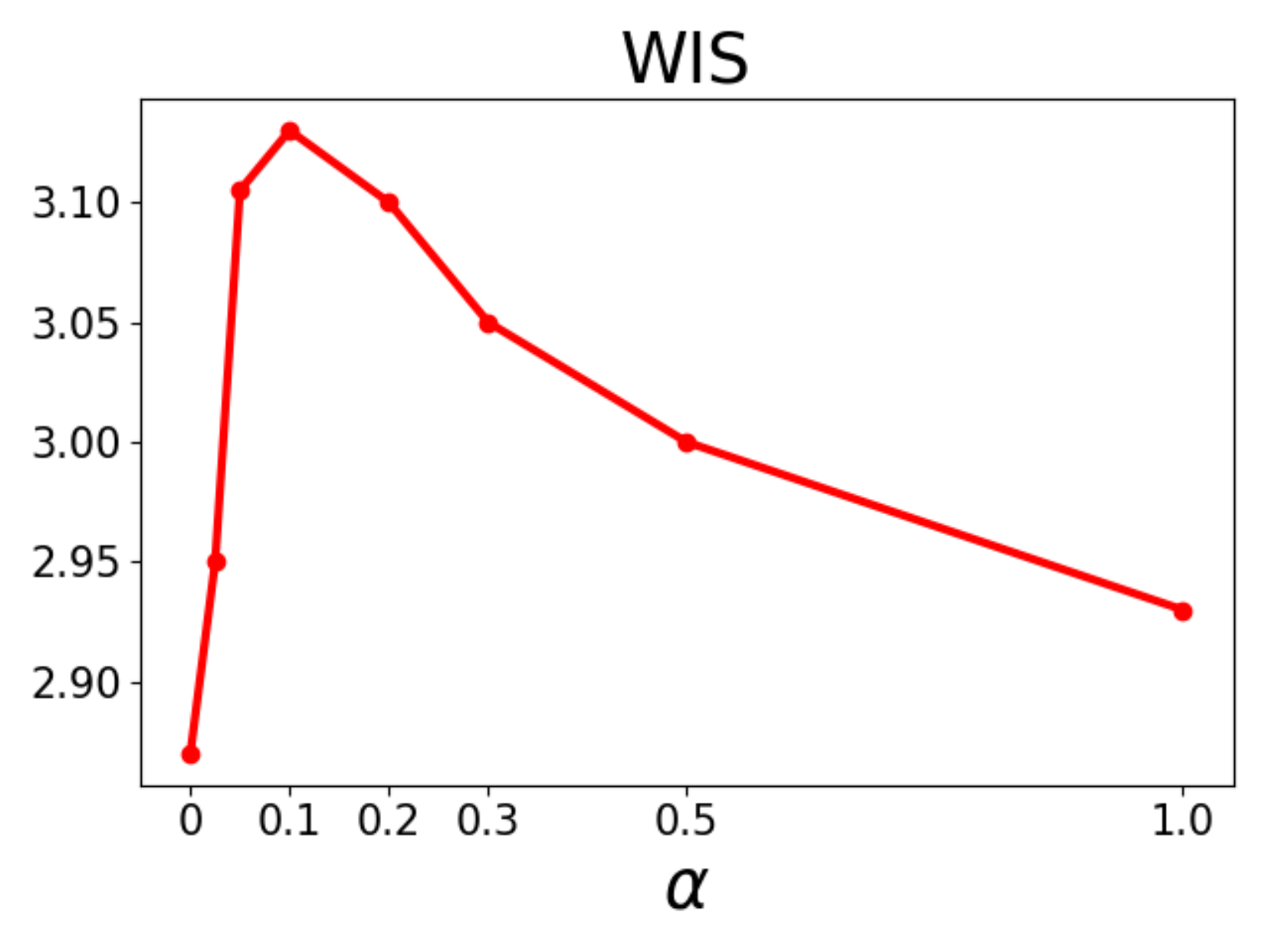}}
\vspace{-4mm}
\caption{Hyper-parameter optimization } 
\label{fig:para}
\end{figure}

\noindent
\textbf{Distribution of Actions}: 
Visualization of the action distribution in the 3-dimensional action space on AmsterdamUMCdb are shown in Figure \ref{fig:distribution_ast}. The results show that the proposed model learned similar policies to clinicians. DAC suggests more actions with the lowest and highest PEEP and FiO2. Besides, the learning policies recommend more frequent lower tidal volume compared to clinician policy.

\noindent
\textbf{Comparison of Clinician and DAC policies}:
We find that the mortality rates are lowest in patients for whom clinicians’ actual treatments matched the actions the learned policies recommend both on MIMIC-III and AmsterdamUMCdb datasets. 
Figure~\ref{fig:dose_diff_ast} shows the relations between mortality rate and mechanical ventilation setting difference on AmsterdamUMCdb.

\noindent
\textbf{Hyper-parameter optimization}: 
Figure \ref{fig:para} shows the optimization of parameter $\alpha$ on MIMIC-III dataset. We find the model performance is not sensitive when $0.05 \leq \alpha \le 0.2$. We set $\alpha=0.1$ when training the DAC model. Because the long-term rewards' value range (i.e., from -15 to +15)  is wider than short-term rewards' value range (i.e., from -1 to +1), the weight of long-term reward is smaller than the weight of short-term reward.

\end{document}